\documentclass[letterpaper]{article} 
\usepackage{aaai25}  
\usepackage{times}  
\usepackage{helvet}  
\usepackage{courier}  
\usepackage[hyphens]{url}  
\usepackage{graphicx} 
\usepackage{natbib}  
\usepackage{caption} 
\usepackage{algorithm}

\usepackage{newfloat}
\usepackage{listings}
\DeclareCaptionStyle{ruled}{labelfont=normalfont,labelsep=colon,strut=off} 
\floatstyle{ruled}
\newfloat{listing}{tb}{lst}{}
\floatname{listing}{Listing}
\usepackage{algorithm}
\usepackage{algpseudocode}
\usepackage{adjustbox}
\usepackage{mathtools} 
\usepackage{amsmath}
\DeclareRobustCommand{\taustar}{\ensuremath{\overset{*}{\tau}}} 
\newcommand{\Lk}{\ensuremath{\mathbf{L}^{-1}_k}}

\title{Deep reinforcement learning with time-scale invariant memory}
\author {
    Md Rysul Kabir,
    James Mochizuki-Freeman,
    Zoran Tiganj
}
\affiliations {
    Department of Computer Science\\
    Luddy School of Informatics, Computing, and Engineering\\
    Indiana University Bloomington\\
    mdrkabir@iu.edu, jmochizu@iu.edu, ztiganj@iu.edu
}

\begin{document}

\maketitle

\begin{abstract}
The ability to estimate temporal relationships is critical for both animals and artificial agents. Cognitive science and neuroscience provide remarkable insights into behavioral and neural aspects of temporal credit assignment. In particular, scale invariance of learning dynamics, observed in behavior and supported by neural data, is one of the key principles that governs animal perception: proportional rescaling of temporal relationships does not alter the overall learning efficiency. Here we integrate a computational neuroscience model of scale invariant memory into deep reinforcement learning (RL) agents. We first provide a theoretical analysis and then demonstrate through experiments that such agents can learn robustly across a wide range of temporal scales, unlike agents built with commonly used recurrent memory architectures such as LSTM. This result illustrates that incorporating computational principles from neuroscience and cognitive science into deep neural networks can enhance adaptability to complex temporal dynamics, mirroring some of the core properties of human learning.
\end{abstract}

\section{Introduction}
Learning temporal relationships between cause and effect is critical for successfully obtaining rewards and avoiding punishments in a natural environment. 
Humans and many other animals can estimate the temporal duration of events and use that estimate as an integral component of decision-making. Furthermore, the ability to do this rapidly and flexibly across a wide range of temporal scales has been critical for survival. While machine learning systems also possess the capacity to represent the elapsed time, typically via recurrent connections, they often struggle with learning temporal relationships, especially when those are extended over multiple scales.

The mammalian ability to estimate time and learn exhibits scale invariance across a wide range of temporal scales, spanning from seconds to several minutes \citep{buhusi2005makes, gibbon1977scalar, buhusi2009interval, balci2020peak, GallGibb00, balsam2009temporal, gallistel2024time}. A scale invariant system has a linear relationship between the mean estimated time and the actual time, with a constant coefficient of variation. In other words, the relative uncertainty or error in time estimation remains constant as the duration increases. This is known as the  Weber-Fechner law \citep{Fech60}, which states that the just noticeable difference between two stimuli is proportional to the magnitude of the stimuli. Hence the ratio between these two quantities is constant, implying that the perceived magnitude is on a logarithmic scale. This law is foundational for understanding mammalian perceptions and spans virtually all perceptual domains except angles \citep{gibbon1977scalar, Wilk15}. Scale invariance in learning is demonstrated in classical conditioning where animals observe a salient stimulus followed by a reward. When the temporal distance between stimulus and reward is rescaled by the same factor as between two stimuli, the number of trials the animal needs to learn the value of the stimulus remains unchanged. While previous studies have demonstrated this effect on the order of about a minute  \citep{GallGibb00, balsam2009temporal}, recent work has demonstrated that it holds for at least 16 minutes \citep{gallistel2024time}. Contrary to biological organisms, the performance of machine learning systems is typically not scale invariant. Machine learning systems tend to perform well only at a limited set of scales and require adjustments of hyperparameters such as learning rate, temporal resolution and temporal discounting to learn problems at different scales.  

A number of neuroscience studies have investigated the neural underpinning of temporal representations in tasks such as interval timing and temporal bisection \citep{emmons2017rodent,matell2004cortico,kim2013neural,kim2017optogenetic,parker2014d1,mello2015scalable,narayanan2016ramping,gouvea2015striatal,tiganj2017sequential,donnelly2015ramping}. Neural activity observed in several brain regions, including the hippocampus, prefrontal cortex and striatum, was often characterized by one of two traits: (1) ramping/decaying activity, where neurons monotonically increase/decrease their firing rate as a function of time following some salient stimulus (such as the onset of the interval timing cue); and (2) sequential activity, where neurons activate sequentially following some salient stimulus, with each neuron elevating its firing rate for a distinct period of time. These sequentially activated neurons, also known as \textit{time cells} (as a temporal analog of \textit{place cells} \citep{o1976place}) have been observed in the hippocampus and the prefrontal cortex in several studies over the recent years \citep{pastalkova2008internally, macdonald2011hippocampal, salz2016time, cruzado2020conjunctive, eichenbaum2014time, macdonald2021crucial}. Sequential neural activity is also scale invariant: the width of the temporal windows in time cells increases with the peak time and the ratio of the peak times of the adjacent cells is constant (geometric progression), indicating uniform spacing along a logarithmic axis  \citep{cao2022internally}. 

In this study, we draw on insights from neuroscience and cognitive science to enable more flexible learning in deep reinforcement learning (RL) agents by incorporating scale invariant temporal memory.
The memory model is based on previous work in computational neuroscience \citep{shankar2012scale,ShanHowa13}, which has been used to explain a broad range of phenomena in neuroscience \citep{howard2014unified}, including the emergence of time and place cells, as well as findings from cognitive psychology memory experiments, such as free recall and judgment of recency \citep{howard2015distributed, tiganj2021computational}. 
We integrate scale invariant memory into RL agents and evaluate their performance across tasks designed to span a wide range of temporal scales. We compare the performance and neural activity of proposed agents to agents built with standard memory architectures, such as LSTM \citep{hochreiter1997long} and simple RNN. Through theoretical analysis and experiments, we demonstrate that agents with scale invariant memory effectively generalize and maintain strong performance across rescaled temporal relationships.

\section{Prior work}
Previous computational work proposed that ramping/decaying activity and time cells provide a representation essential for timing and learning temporal relationships \citep{balci2016decision, howard2014unified}. This has also been studied in the RL context \citep{petter2018integrating, namboodiri2022real,ludvig2008stimulus}. A number of computational neuroscience studies have closely examined and attempted to model neural activity during interval timing and similar time-related tasks \citep{grossberg1989neural,wang2018flexible,jazayeri2015neural,cueva2020low,perez2018synaptic,raphan2019modeling}. Together, these models provide remarkable insights into neural mechanisms of temporal learning from computational neuroscience and cognitive science perspectives. Our efforts focus on examining temporal learning in artificial RL agents trained using error backpropagation. We believe this effort complements computational neuroscience efforts for building neural models of time-scale invariant learning since we specifically evaluate agents based on a computational neuroscience model of memory. 

\citet{deverett2019interval} studied interval reproduction in deep RL agents and provided valuable insights about the capabilities of these agents. Our work extends this approach, as we closely examine the neural activity of the artificial agents and conduct experiments at different temporal scales. 
Previous work has also shown that under particular circumstances (mnemonic demand), neural activity in deep RL agents resembles the activity of time cells \citep{lin2021time}. Other types of brain-like neural activity have also been observed in deep learning systems \citep{sorscher2019unified,schaeffer2022no}, which in some cases led to improved performance of deep RL agents \citep{banino2018vector}. Furthermore, recent computational models of hippocampal activity have been linked to modern deep learning architectures, such as transformers \citep{whittington2021relating}. Our work complements these efforts in brain-inspired AI by comparing neural activity in biological and artificial systems and evaluating whether systems whose activity is closer to biological counterparts indeed perform better.  

Neural networks that use temporal convolution to construct an offline version of scale invariant memory have been proposed in \citet{jacques2021deepsith,jacques2022deep}, but these have not been explored in the context of online temporal learning or  deep RL. An approach similar to ours has been used to develop systems with scale invariant temporal discounting \citep{TanoEtal20,masset2023multi,MomeHowa18,tiganj2019estimating}. A similar approach has also been used to build systems that can learn to represent variables such as numerosity and position \citep{maini2023representing, mochizuki2024incorporating}. These approaches are complementary to ours since we focus on scale invariance in temporal memory of deep RL agents.

\section{Environments}

\begin{figure*}[h!]
	\centering
    \begin{tabular}{ll}
        \textbf{A} &
        \textbf{B} \\
        \includegraphics[width=0.34\textwidth]{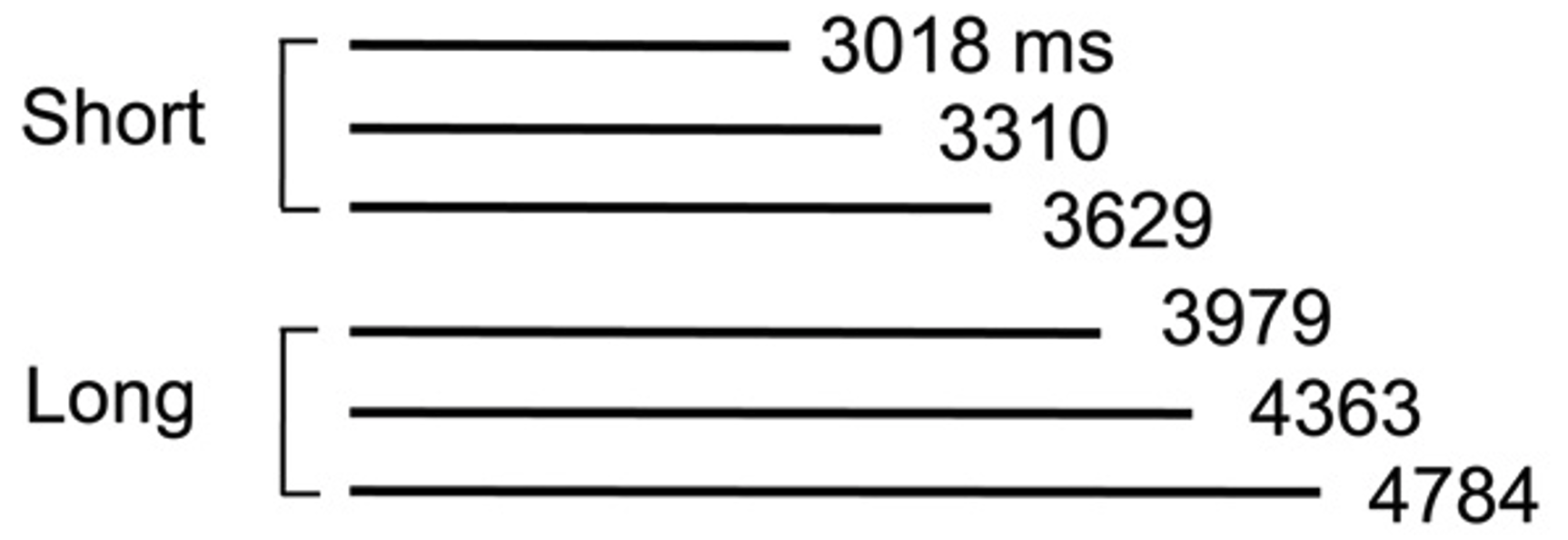} &
        \includegraphics[width=0.60\textwidth]{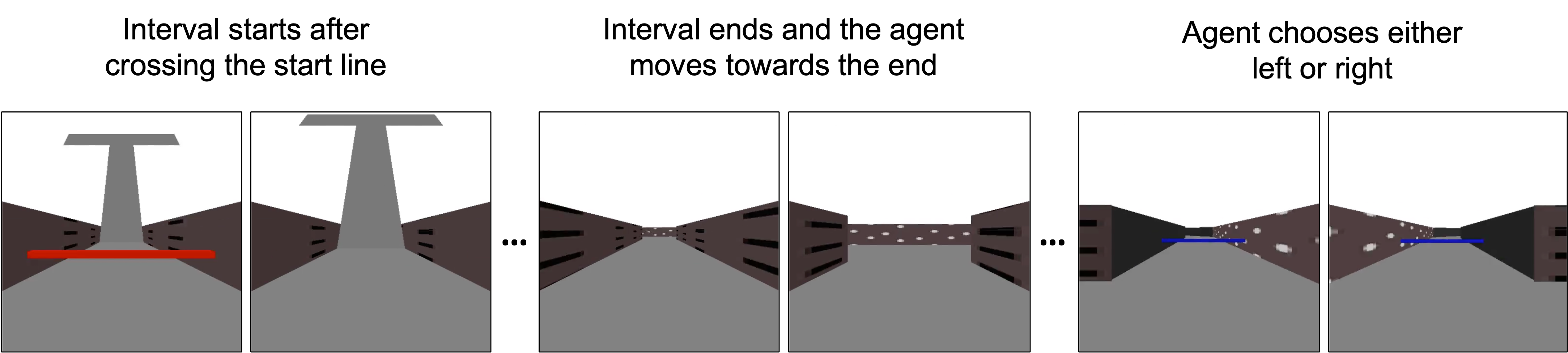} \\
    \end{tabular}
	\caption{\textbf{A.} The six temporal intervals used in the task. At each trial, a random interval is selected and the agent has to indicate whether the interval was long or short. \textbf{B.} Schematic of the environment. After the agent crosses the start line, one of six delay intervals is presented.  \label{fig:env}}
\end{figure*}

\subsection{Interval timing}
The interval timing environment was inspired by a neuroscience study where rats had to discriminate between two groups of intervals, long and short, each having three possible durations \citep{kim2013neural} (Fig.~\ref{fig:env}A). The interval begins when the agent crosses the start line (indicated by the red line in Fig.~\ref{fig:env}B). The duration of the interval is randomly chosen from six equally probable durations. After the end of the interval period, the central bridge of the T-maze drops, and the agent freely navigates towards the end of the track. When the agent reaches the end, it has to choose one of the goal locations (indicated by the blue lines in Fig.~\ref{fig:env}B) depending on whether the interval was from the long or short group of intervals. We used two versions of this environment: one with 3D realistic visual observations developed using the open-source PyBullet physics engine \citep{coumans2021}, and the other that consists of a 1D observation space. The purpose of the simple 1D environment was to facilitate interpretability of the results, while the 3D realistic environment was used to evaluate the scalability of the proposed approach.

In the 3D environment, at each time step, the agent receives a three-dimensional pixel observation of shape $60 \times 60 \times 3$ and performs any of the three possible actions: \textit{left}, \textit{right} and \textit{forward}. The \textit{forward} action moves the agent straight towards the end of the track at a fixed speed of $1$cm/step. Each \textit{left} and \textit{right} action realigns the agent in $7.5^{\circ} $ increments to a max/min of $\pm15^{\circ} $ along the straight portion of the T-maze and without a bound at the end. Once the agent reaches the goal positions, it receives a reward of either $10$ for a correct or $0$ for an incorrect choice. It also receives a reward of $-0.1$ if it tries to turn beyond $15^{\circ} $ in either direction when on the straight track, or if it collides with the back wall of the environment. 

In the simple version of the environment (top row in Fig.~\ref{fig:Cog-RNN}B), the observation space is one-dimensional such that a $\delta$ pulse is introduced after the fixation period to signal the beginning of an interval, followed by a second $\delta$ pulse that marks the end of the interval. The interval is followed by a delay period, after which the agent chooses either a ``left'' or ``right'' action, similar to the 3D environment. In this and subsequent environments, the agent receives a reward of $1.0$ for a correct choice and $-1.0$ for an incorrect choice. 

\subsection{Interval discrimination}
This simple environment was based on a duration discrimination task \citep{intervaldisc}. In each trial, two stimuli are presented sequentially with a delay period separating them. The agent is required to determine which of the two stimuli has a longer duration. The observation space is one-dimensional. Consistent with the 1D interval timing task, the onset and termination of each interval are signaled by a $\delta$ pulse.

\subsection{Delayed-match-to-sample}
We used an existing delayed-match-to-sample environment from NeuroGym \citep{molano2022neurogym}. This environment was inspired by a common memory task \citep{dms} in which a sample stimulus is followed by a delay period and a test stimulus. To receive a reward, the participant needs to determine whether the test stimulus matches the sample stimulus. The observation space for this environment is a vector comprising three features.  

\subsection{Interval reproduction}
This simple environment is inspired by the interval reproduction task described by \citet{deverett2019interval}. In each trial, two stimuli are presented sequentially, separated by an interval that the agent must later reproduce. During the reproduction phase, which occurs after the second stimulus, the agent must perform an action to replicate the interval within a $20\%$ tolerance to receive a reward. In line with the other 1D tasks, the onset and termination of the interval are marked by a $\delta$ pulse.

\section{Model}

\begin{figure}[h!]
	\centering
        \includegraphics[width=1\columnwidth]{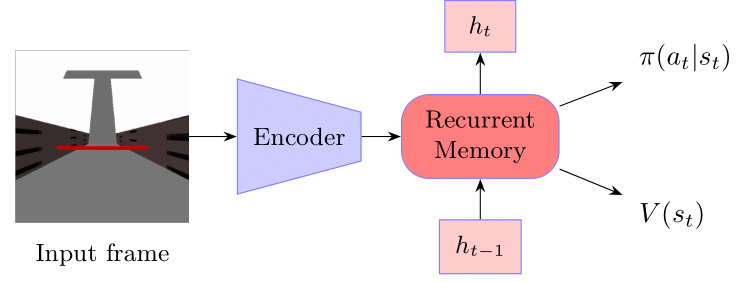} 
	\caption{Architecture of the (RL) agent. Observations from the environment are processed by a convolutional neural network to extract feature representations. These features are then passed to a recurrent memory module (simple RNN, LSTM or CogRNN), which captures temporal dependencies and provides context for the policy network ($\pi$) and value network ($V$).}
        \label{fig:agent}
\end{figure}

Our recurrent deep learning architecture consists of three major parts (Fig.~\ref{fig:agent}): \textit{encoder} (used only for the 3D interval timing environment), \textit{core}, and \textit{agent}. The \textit{encoder} consists of three consecutive convolutional layers. The resulting output from the last convolutional layer is flattened and passed through a fully connected layer. This encoded representation is then fed into the \textit{core}, the recurrent memory of our architecture. We used three different kinds of recurrent cores: RNN, LSTM, and the cognitively inspired RNN, described in detail below, which we abbreviate as CogRNN. Lastly, the output from the \textit{core} is passed through two attention layers (for the 3D interval timing environment) or a dense layer (for other environments unless specified otherwise). This output is then fed into the agent part of the network, which has two branches: (i) policy network and (ii) value network. At every time step, the agent chooses an action depending on the output of the policy network. Outputs from the policy and value networks are used to calculate losses, which are then used in backpropagation for gradient-based parameter updates. 

\begin{figure*}[h!]
    \centering
    \begin{tabular}{lll}
    \textbf{A} &
    \textbf{B} &
    \textbf{C} \\
        \adjustbox{valign=t}{\includegraphics[width=0.31\textwidth]{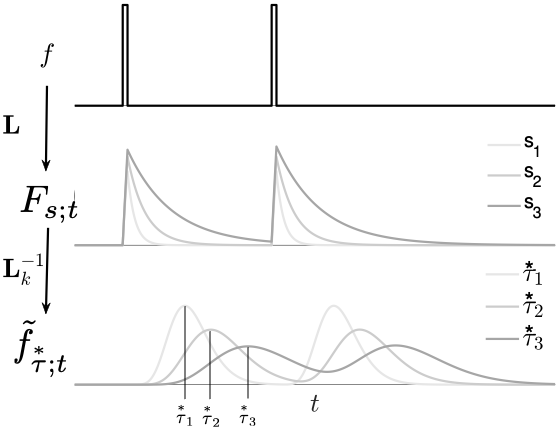}} &
        \adjustbox{valign=t}{\includegraphics[width=0.31\textwidth]{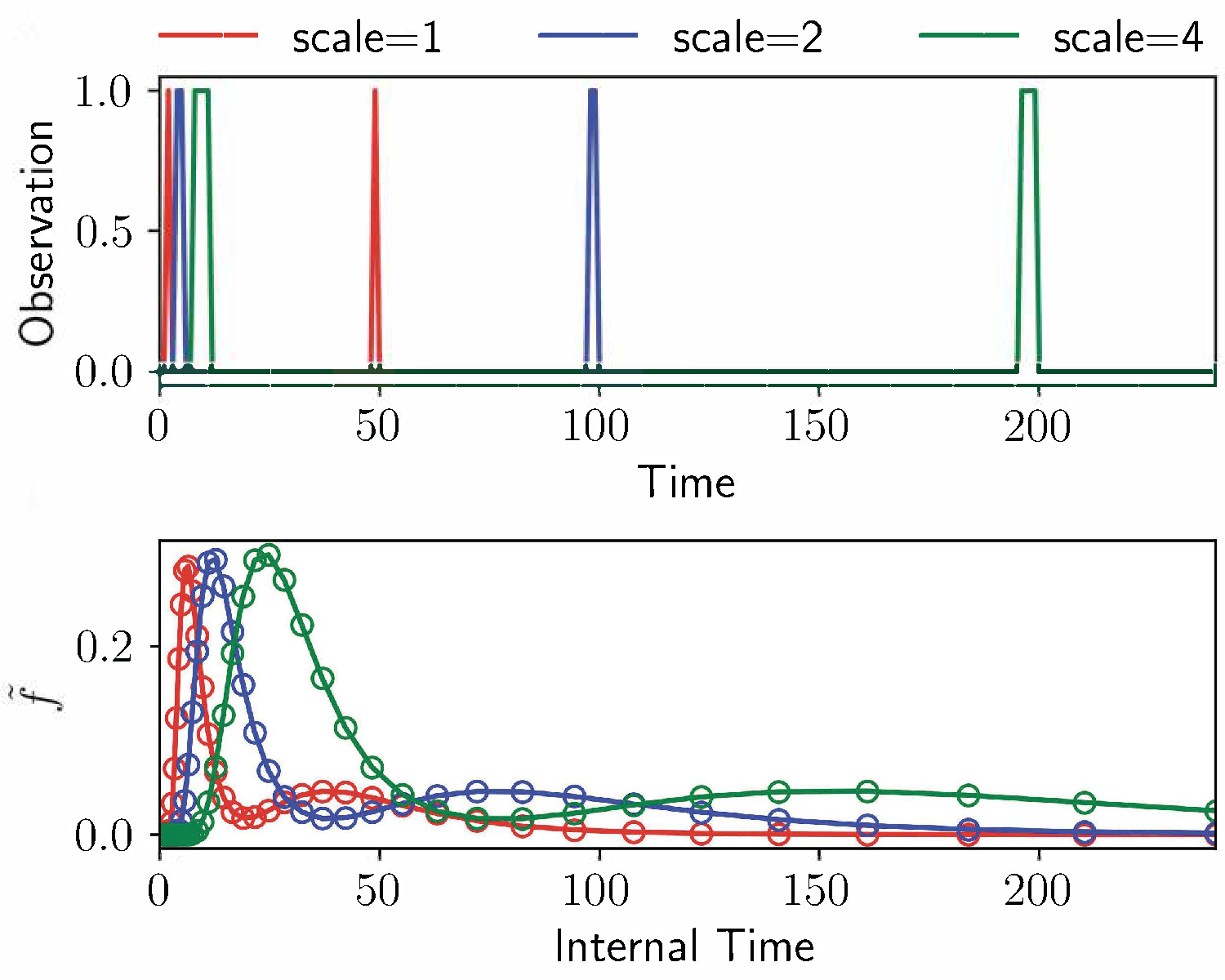}} &
        \adjustbox{valign=t}{\includegraphics[width=0.31\textwidth]{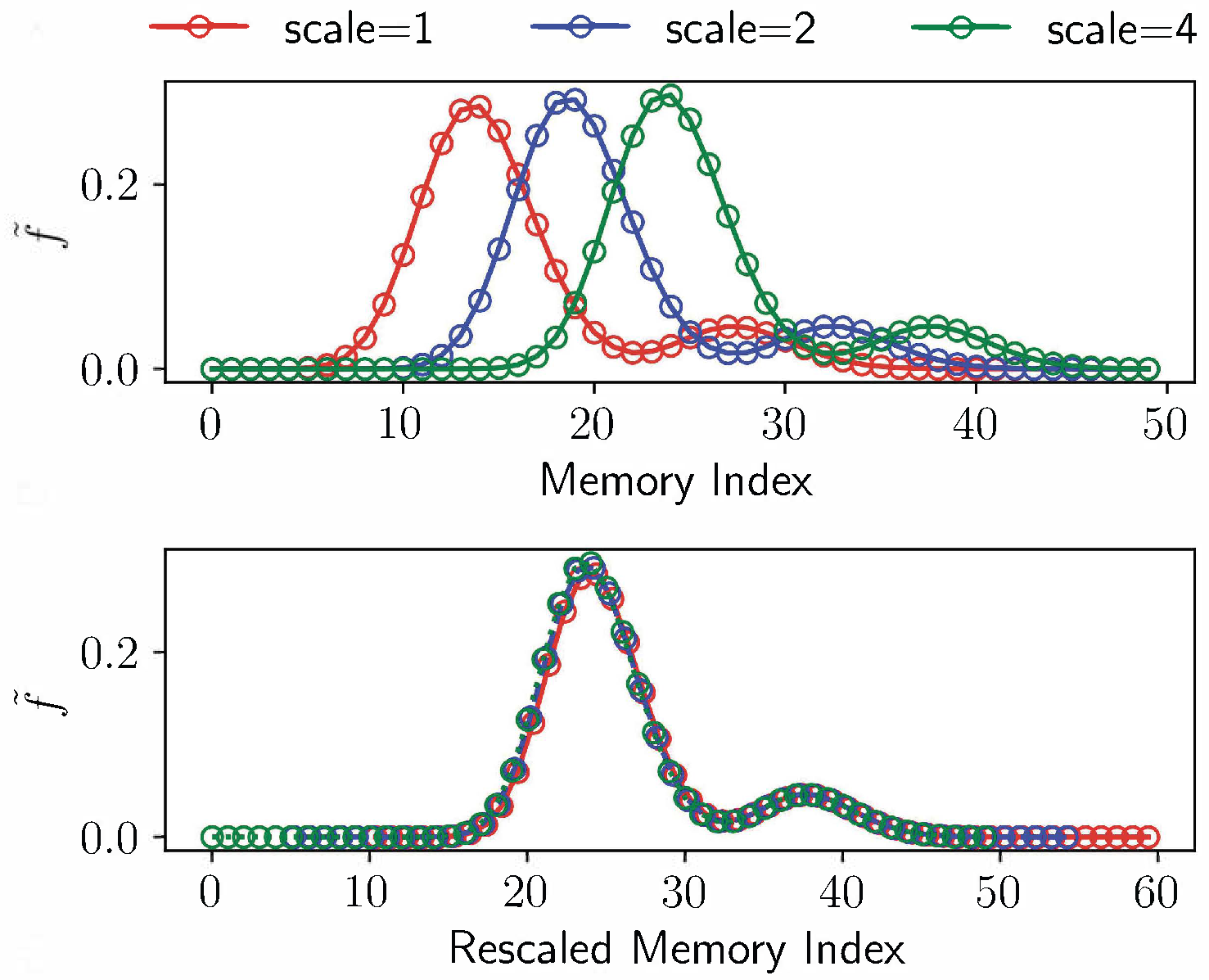}} 
    \end{tabular}
    \caption{\textbf{A.} Response of the CogRNN to $\delta$ pulses. Neurons in $F_{s;t}$ decay exponentially at a spectrum of time constants $\mathbf{s}$ implementing a discrete approximation of a real-domain Laplace transform. Neurons in $\tilde{f}_{\taustar;s}$ activate sequentially, resembling time cells. \textbf{B.} Log-compressed memory (bottom) of three  signals that are rescaled versions of each other (top) at time $t=250$. Each circle represents the activity of individual neurons and their position along the x-axis corresponds to their peak time. \textbf{C.} Log-compressed memory turns rescaling into translation. The top plot is the same as the bottom plot in \textit{B}, but with the x-axis corresponding to the neuron index instead of the peak time. \label{fig:Cog-RNN}}
\end{figure*}

\subsection{Scale invariant memory network}

Building on models from computational and cognitive neuroscience \citep{shankar2012scale,howard2014unified}, we designed a neural network architecture that maintains a scale invariant memory. Specifically, this network constructs an approximation of a real-domain Laplace transform of the temporal history of the input signal and then constructs an approximate inverse of the history -- giving rise to an internal timeline of the past along a log-compressed axis characterized by sequentially activated neurons (Fig.~\ref{fig:Cog-RNN}A). 

To construct a scale invariant memory we use a network composed of two layers. The input coming from the \textit{encoder}, which we label as $f$, is fed into a recurrent layer ($F$) with the weights analytically computed to approximate the real-domain Laplace transform of the temporal history of the input. The output of the recurrent layer is mapped through a linear layer with analytically computed weights implementing the inverse Laplace transform $\tilde{f}$ (Fig.~\ref{fig:Cog-RNN}A). Below we describe this procedure step by step, first the continuous-time formulation and then discrete, neural network implementation.

\subsubsection{Continuous-time formulation}

Given a one-dimensional input signal $f(t)$, we define a modified version of the Laplace transform $F(s;t)$:
\begin{equation}
	F(s;t) = \int_{0}^t e^{-s\left(t - t' \right)}f(t') dt'.
		\label{eq:Laplace_int}
\end{equation}
This modified version differs from the standard Laplace transform only in the variable $s$. Instead of $s$ being a complex value composed of real and imaginary parts, we restrict $s$ to a positive real value. This modification simplifies the neural network implementation while giving us the computational benefits of the standard Laplace transform, as illustrated below. Note that $F$ is also a function of time $t$. This implies that at every moment, we construct the Laplace transform of the input function up to time $t$: ${f(0 \leq t'<t) \xrightarrow{L} F(s;t)}$. 

To construct the temporal history of the input, we need to invert the Laplace transform. The inverse, which we denote as $\tilde{f}(\taustar;t)$, can be computed using Post's inversion formula \citep{Post30}:  
\begin{equation}
	\tilde{f}(\taustar;t) = \Lk F(s;t) = \frac{(-1)^k}{k!} s^{k+1} \frac{d^k}{ds^k}F(s;t), \label{eq:Laplace_inv}
\end{equation}
where $\taustar := k/s$ and $k \rightarrow \infty$ (see \citet{TanoEtal20,horvath2020numerical} for alternative approaches to computing the inverse transform).

\subsubsection{Neural networks implementation}

To describe a neural network approximation of the Laplace transform, we first rewrite Eq.~\ref{eq:Laplace_int} in a differential form:
\begin{equation}
	\frac{dF(s;t)}{dt} = -s F(s;t) + f(t).  \label{eq:Laplace_diff}
\end{equation}
The impulse response (response to input $f(t) = \delta(0)$) of $F(s;t)$ decays exponentially as a function of time $t$ with decay rate $s$: $e^{-st}$ (the second row in Fig.~\ref{fig:Cog-RNN}A). Note that this is a linear transform, so $F(s;t)$ will be a convolution between $f(t)$ and the impulse response. 

We implement an approximation of the modified Laplace and inverse Laplace transform as a two-layer neural network with analytically computed weights. The first layer implements the modified Laplace transform through an RNN. The second layer implements the inverse Laplace transform as a dense layer with weights analytically calculated to compute a $k$-th order derivative with respect to $s$. 

While in the Laplace domain $s$ is a continuous variable, here we redefine $s$ as a vector of $N$ elements. We can now write a discrete-time approximation of Eq.~\ref{eq:Laplace_diff} as an RNN with a diagonal connectivity matrix: 
\begin{equation}
	    F_{s;t} = \mathbf{L} F_{s;t-1} + f_t  \label{eq:Laplace_discrete}, 
\end{equation}
where $\mathbf{L}$ is an $N \times N$ recurrent matrix $\mathbf{L} := e^{-\mathbf{S}\Delta t}$ implementing the discrete Laplace transform operator and $\mathbf{S}$ is a diagonal matrix composed of $s$ values. At every time step $t$, $F_{s;t}$ is an $N$-element vector. For brevity of notation, we assume that the duration of a discrete-time step $\Delta t = 1$. 
Following Eq.~\ref{eq:Laplace_inv}, a discrete approximation of the inverse Laplace transform, $\tilde{f}_{\taustar;t}$, can be implemented as a dense layer on top of $F_{s;t}$. The connectivity matrix of the dense layer is \Lk (see \citet{maini2023representing} for the derivation of the exact matrix form of \Lk).

To interpret $\tilde{f}_{\taustar;t}$ and to select $s$ values in an informed way, we compute the impulse response of $\tilde{f}_{\taustar;t}$. For input $f(t) = \delta(0)$, the activity of $\tilde{f}_{\taustar;t}$ is: 
\begin{equation}
	     \tilde{f}_{\taustar;t} = \frac{1}{t} \frac{k^{k+1}}{k!} \left(\frac{t}{\taustar}\right)^{k+1} e^{-k\frac{t}{\taustar}}. \label{eq:Laplace_inv_discrete_impulse}
\end{equation}

The impulse responses of units in $\tilde{f}_{\taustar;t}$ is a set of unimodal basis functions (Fig.~\ref{fig:Cog-RNN}A). To better characterize their properties, we first find the peak time by taking a partial derivative with respect to $t$, equate it with 0 and solve for $t$: $\partial \tilde{f}_{\taustar;t}/ \partial t = 0 \rightarrow t=\taustar$. 
Therefore, each unit in $\tilde{f}_{\taustar;t}$ peaks at $\taustar$. 

To further characterize our approximation, we express the width of the unimodal basis functions of the impulse response of $\tilde{f}_{\taustar;t}$ through the coefficient of variation $c$: $c=1/\sqrt{k+1}$. Importantly, $c$ does not depend on $t$ and $\taustar$, implying that the width of the unimodal basis functions increases linearly with their peak time. Therefore, when observed as a function of $\log(t)$, the width of the unimodal basis functions is constant. 

We choose values of $\taustar$ as log-spaced between some minimum $\taustar_{min}$ and maximum $\taustar_{max}$ as specified in the next section. Note that fixing the values of $\taustar$ and choosing $k$ also fixes values of $s$ since $s=k/\taustar$, so $s$ is not a trainable parameter. Because of the log-spacing and because $c$ does not depend on $t$ and $\taustar$, when analyzed as a function of $\log(t)$, the unimodal basis functions are equidistant and equally wide, providing uniform support over the $\log(t)$ axis (Fig~\ref{fig:Cog-RNN}C). Because the described system is linear, any input function is represented as a convolution with a set of these basis functions. As we demonstrate next, this produces a log-compressed memory that is scale invariant.

\subsubsection{Invariance to temporal rescaling}
Following Eq.~\eqref{eq:Laplace_inv} and Eq.~\eqref{eq:Laplace_inv_discrete_impulse} we note that $\tilde{f}(\taustar;t)$ is scale invariant in the sense that rescaling $\tilde{f}(\taustar;t) \longrightarrow \tilde{f}(\taustar;a t)$ can be undone by setting $\taustar_i \longrightarrow \taustar_i /a$. Choosing $\taustar$ to be log-spaced ($\taustar_i = (1+c)^{i-1} \taustar_{min}$, with $c>0$) makes the rescaling of $\taustar$ equivalent to translation: $\taustar_i  = \taustar_{i+\Delta}$ where $\Delta = log_{1+c}a$. This implies that temporal rescaling will cause a translation of the sequentially activated units (Fig.~\ref{fig:Cog-RNN}B,C).

This approach can be used to either build agents whose temporal memory representation covaries with the temporal scale (i.e., temporal rescaling is converted into translation as described above) or agents with temporal memory that is invariant of the temporal scale (i.e., if time in the environment rescales, the agent will have identical memory representation). When the objective is to build agents that are invariant to temporal rescaling, we apply convolution and pooling over $\tilde{f}$. The output of convolution and pooling is translation invariant, making the network invariant to temporal rescaling (top row in Fig.~\ref{fig:results_invariance}A). We note that since rescaling is converted into translation, the presence of edge effects impacts the output of convolution and pooling, thereby making the invariance imperfect. 

\subsubsection{Intuition}
To provide an intuitive understanding of the scale invariant memory, we refer to Fig.~\ref{fig:Cog-RNN}B. The top row shows three input signals each composed of two $\delta$ pulses. Each of the three signals is a rescaled version of another, i.e. $f_1(t) = f_2(a_1 t) = f_3(a_2 t)$ where $a_i$ is the scaling factor ($a_1=2$ and $a_2=4$ in our example). The second row shows activity of neurons in the scale invariant memory of the above signals (this is the state of the memory at time $t=250$). Neurons in a scale invariant memory are activated sequentially as a function of internal representation of time. Each circle in the plot represents activity of a single neuron and the x-axis corresponds to the log-spaced peak times (hence we refer to it as an internal representation time). Together, these neurons code a  log-compressed memory of the input. This  memory representation converts functions of time $t$ into functions of $log(t)$. If time is rescaled by factor $a$, the resulting representation will be shifted (translated) by $\log(a)$: $f(at) \rightarrow f(\log(at)) = f(\log(a) + \log(t))$. Therefore, the memory representation is covariant with respect to rescaling. The strength of such a representation is that it makes computational problems that occur at different scales equally difficult.
When plotted as a function of neuron index, rather than peak time (top row in Fig.~\ref{fig:Cog-RNN}C), the three signals corresponding to three temporal scales are now translated, rather than rescaled, versions of one another. If the x-axis is shifted by $\log(a)$, the three signals overlap (bottom row in Fig.~\ref{fig:Cog-RNN}C).

\subsection{RL Agent}
We used deep RL agents based on a synchronous version of A3C \citep{mnih2016asynchronous} that uses advantage to calculate gradient-based policy updates. However, as the direct calculation of advantage may lead to high variance and longer training times, we have used an exponentially weighted estimator of advantage called generalized advantage estimator (GAE) \citep{SchulmanMLJA15}.

\subsection{Hyperparameters and training}
Our setup for the 3D interval timing environment used the three following convolutional layers: 32 kernels of size $8 \times 8$ (stride 2), 16 kernels of size $4 \times 4$ (stride 1), and 32 kernels of size $8 \times 8$ (stride 2). The fully connected layer after the convolution layer has 64 nodes. Outputs of all layers had ReLU activation. Following the encoder, the RL agent had either an LSTM network with $256$ hidden units, or the CogRNN architecture with $8$ log-spaced units having $\taustar_{min} = 1$, $\taustar_{max} = 1000$, and $k = 8$. We also introduced two layered multihead-attention \citep{attention} over the $\taustar$ with $8$ heads and $d_{model}$ $128$. Parameters related to the RL algorithm were a discount factor ($\gamma$) of 0.98 and a decay ($\lambda$) factor of 0.95. We used the Adam optimizer with $\beta_{1} = 0.9$, $\beta_{2} = 0.999$, and $\epsilon = 1e^{-8}$.  We also trained the RL agents with varying learning rates, including $0.001$, $0.0001$ and $0.00001$, and selected the best-performing learning rates for each agent. We explored the impact of the hyperparameters (learning rate and entropy coefficient) and provided results as a part of the Supplemental Information. 

\sloppy
For simple environments, we used LSTM networks with $128$ hidden size and the same CogRNN network without the attention network. We updated the model parameters after each trial instead of backpropagating the gradients after horizon number of steps.\footnote{Our implementation is  available in\\
\url{https://github.com/cogneuroai/RL-with-scale-invariant-memory}.}

\begin{figure*}[ht!]
    \centering
    
    \rotatebox[origin=c]{90}{\textbf{ \ }}
    \hfill
    \begin{minipage}{0.23\textwidth}
        \centering
        Interval timing (1D)
    \end{minipage}
    \begin{minipage}{0.23\textwidth}
        \centering
        Interval discrimination (1D)
    \end{minipage}
    \begin{minipage}{0.23\textwidth}
        \centering
        Delayed match to sample (1D)
    \end{minipage}
    \begin{minipage}{0.23\textwidth}
        \centering
        Interval timing (3D) 
    \end{minipage}
    
    
    \rotatebox[origin=c]{90}{CogRNN}
    \hspace{2pt}
    \begin{minipage}{0.23\textwidth}
        \includegraphics[width=\linewidth]{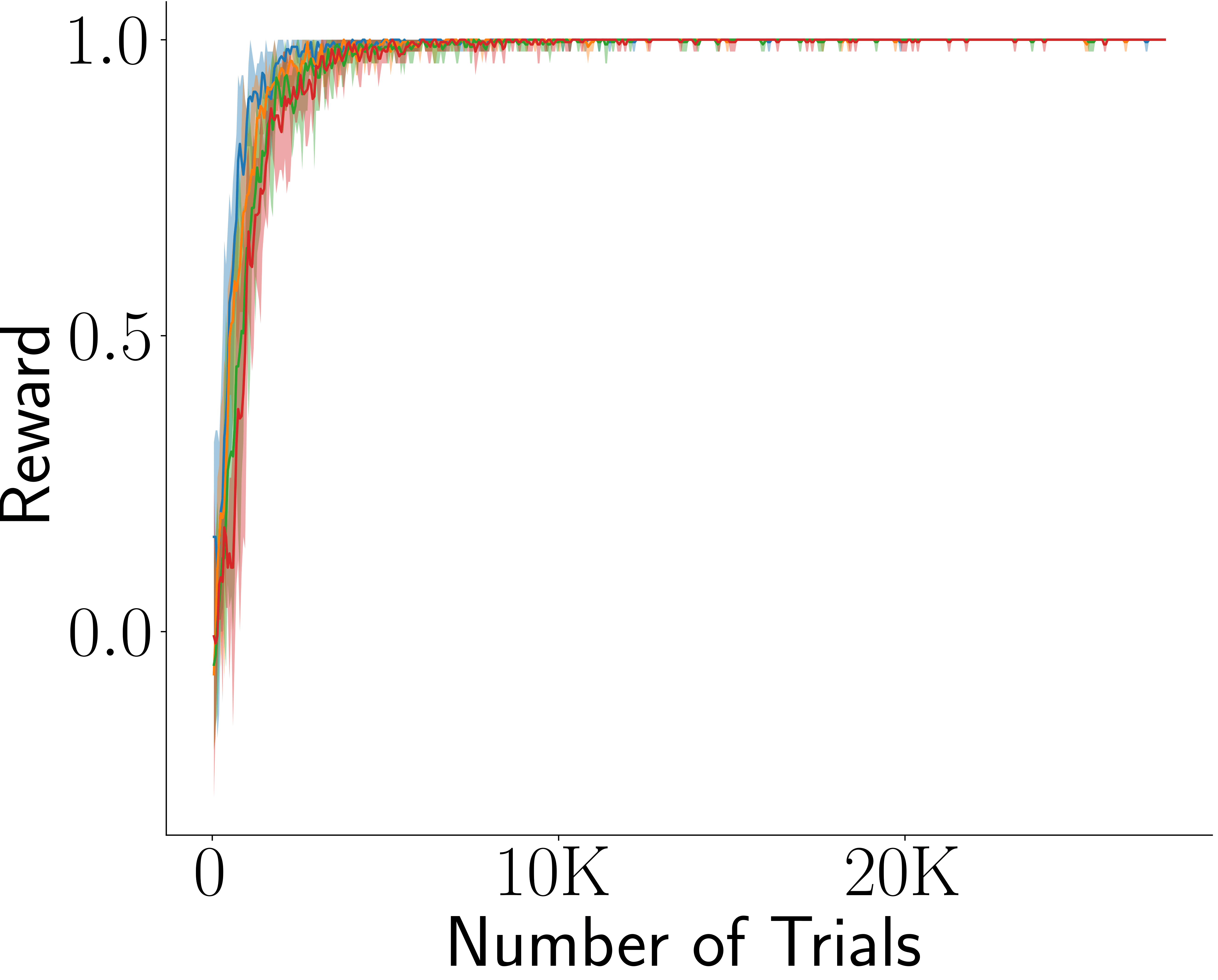} 
    \end{minipage}
    \begin{minipage}{0.23\textwidth}
        \includegraphics[width=\linewidth]{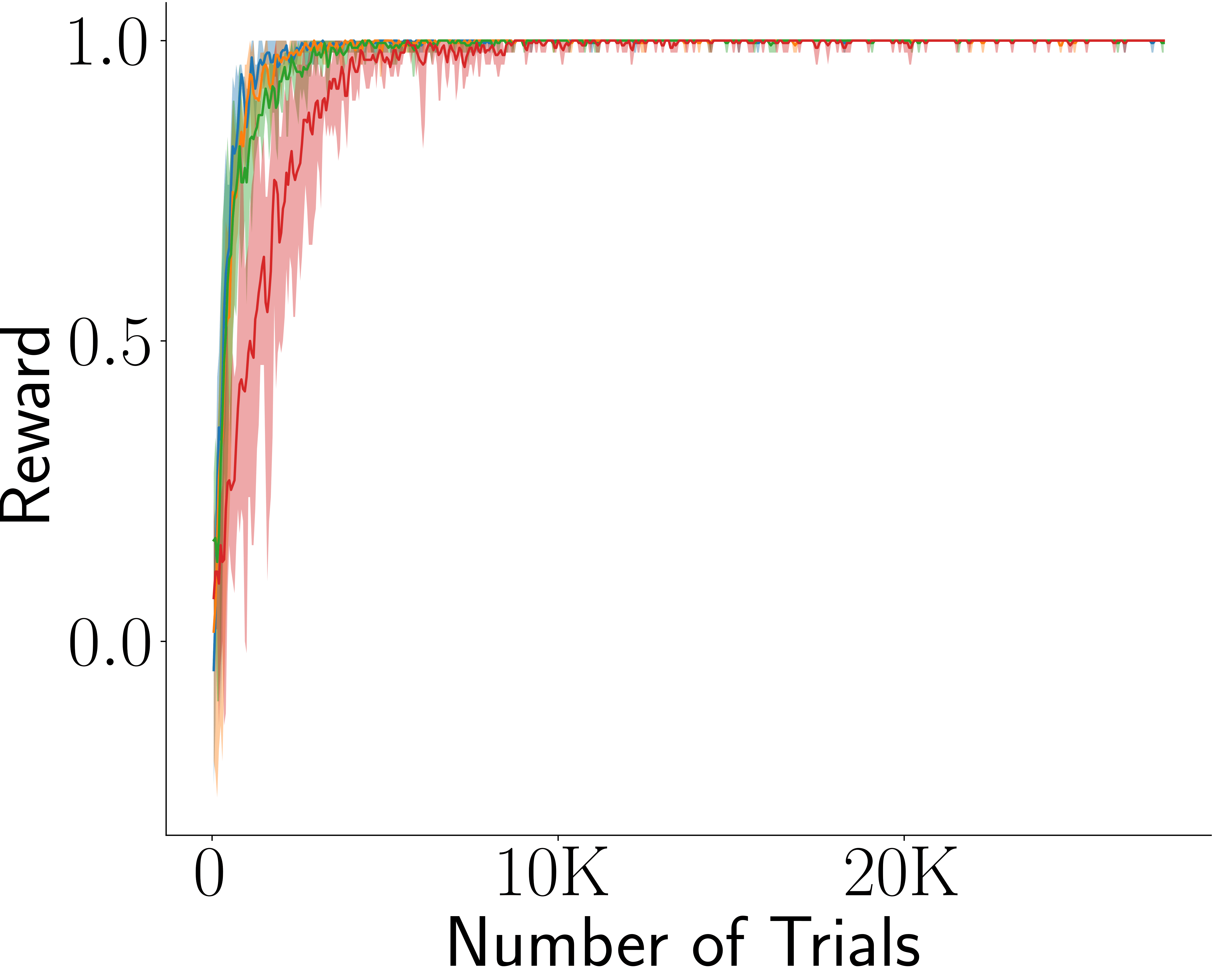} 
    \end{minipage}
    \begin{minipage}{0.23\textwidth}
        \includegraphics[width=\linewidth]{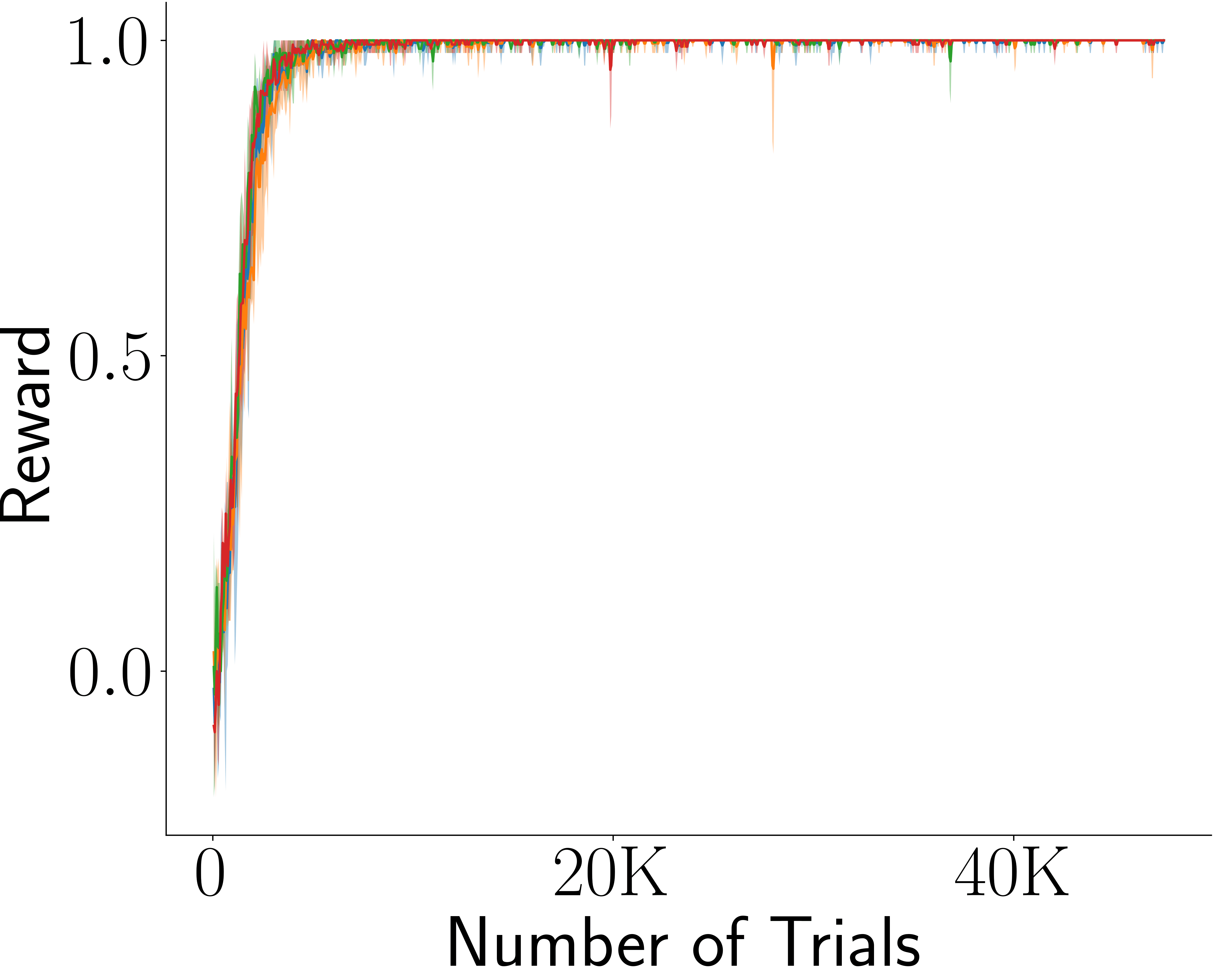} 
    \end{minipage}
    \begin{minipage}{0.23\textwidth}
        \includegraphics[width=\linewidth]{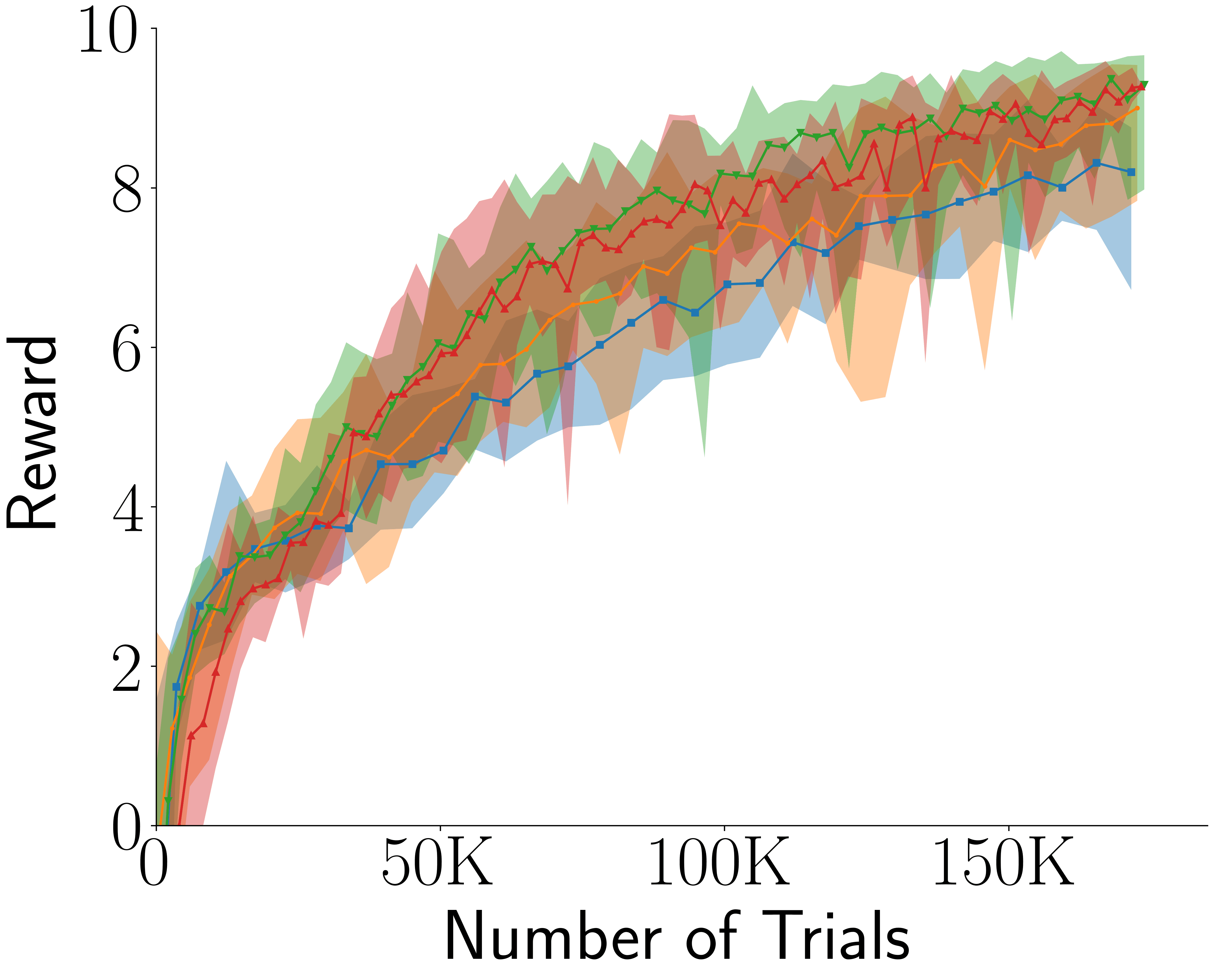}
    \end{minipage}
    
    
    \rotatebox[origin=c]{90}{LSTM}
    \hspace{2pt}
    \begin{minipage}{0.23\textwidth}
        \includegraphics[width=\linewidth]{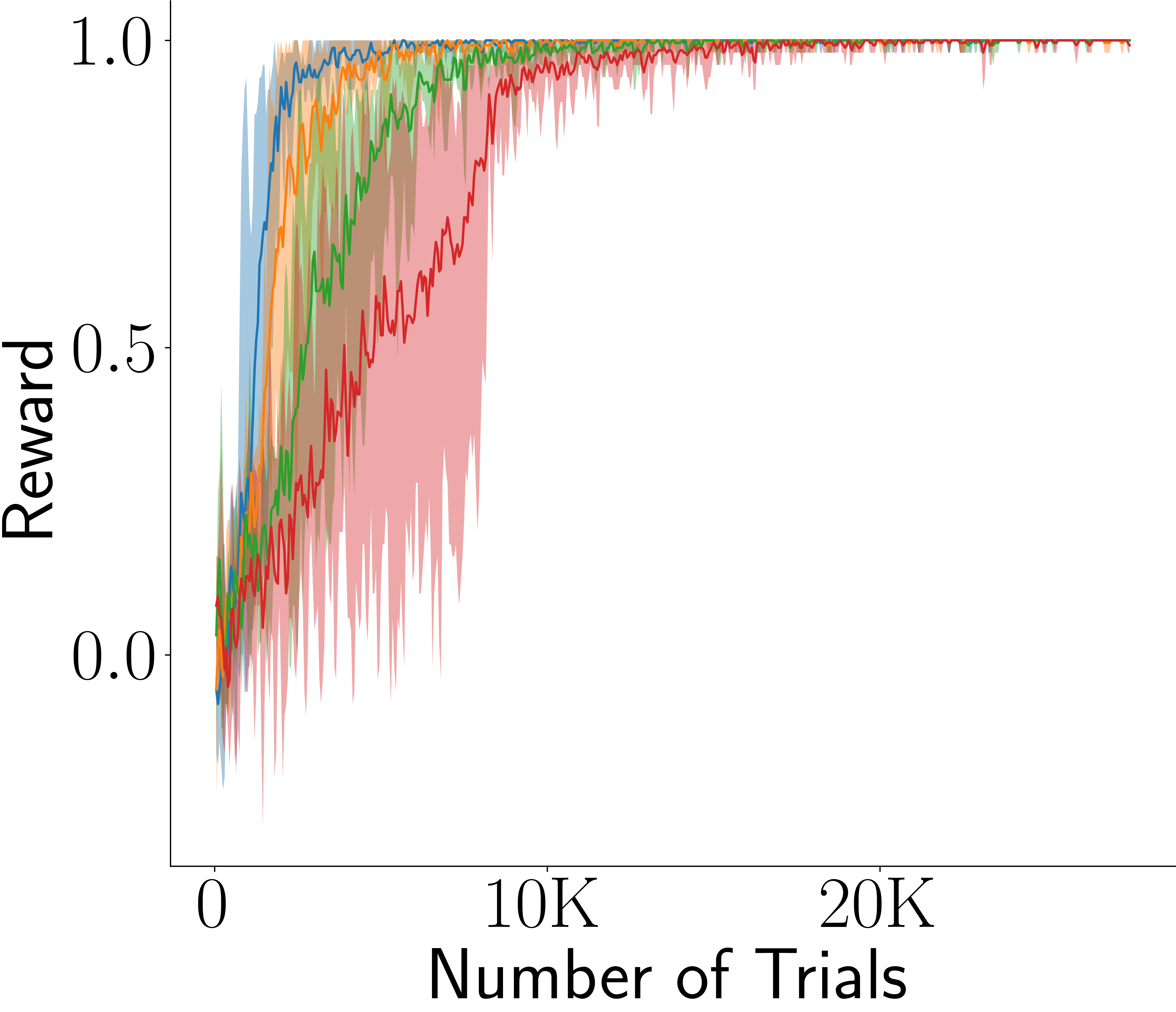} 
    \end{minipage}
    \begin{minipage}{0.23\textwidth}
        \includegraphics[width=\linewidth]{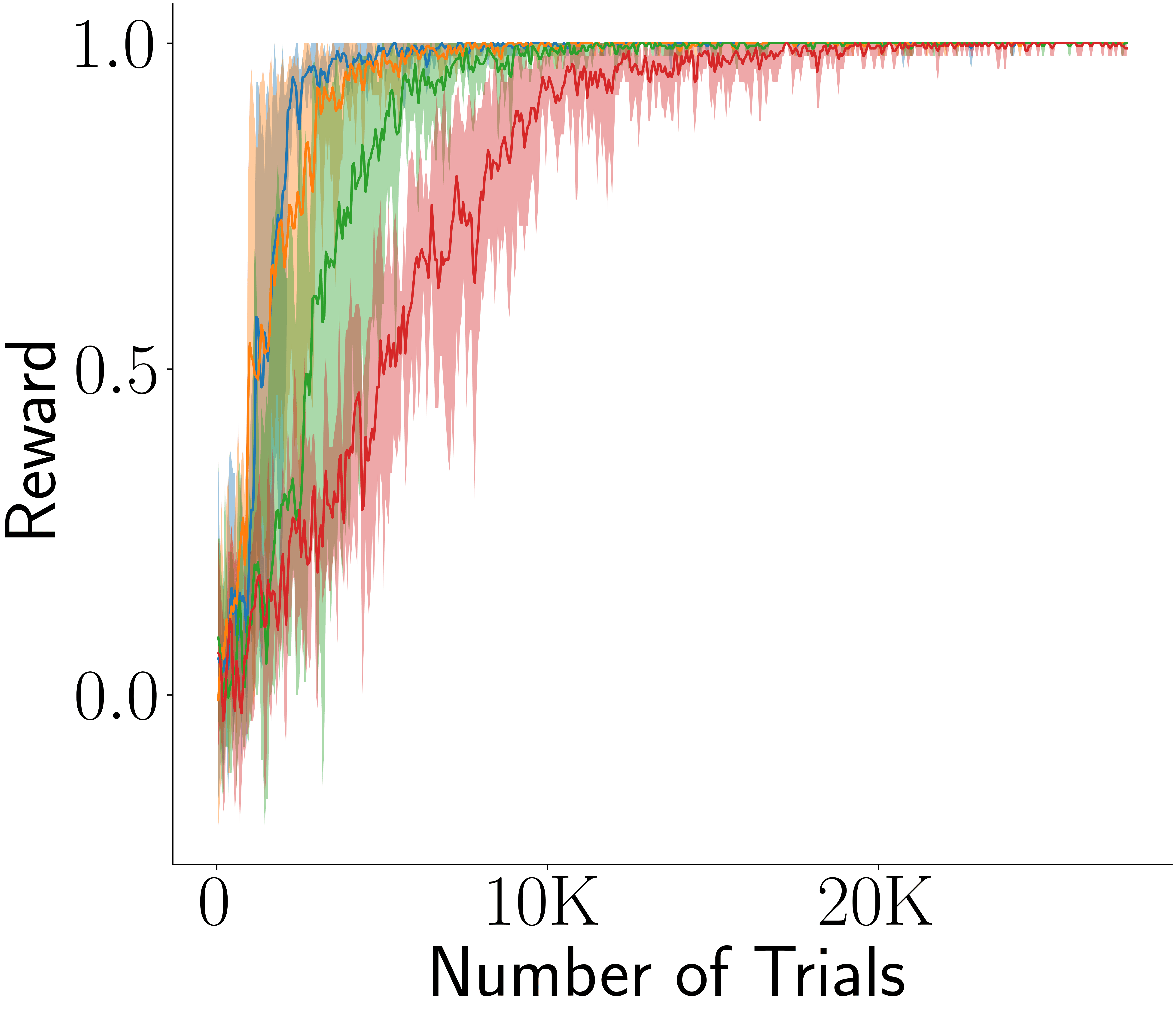} 
    \end{minipage}
    \begin{minipage}{0.23\textwidth}
        \includegraphics[width=\linewidth]{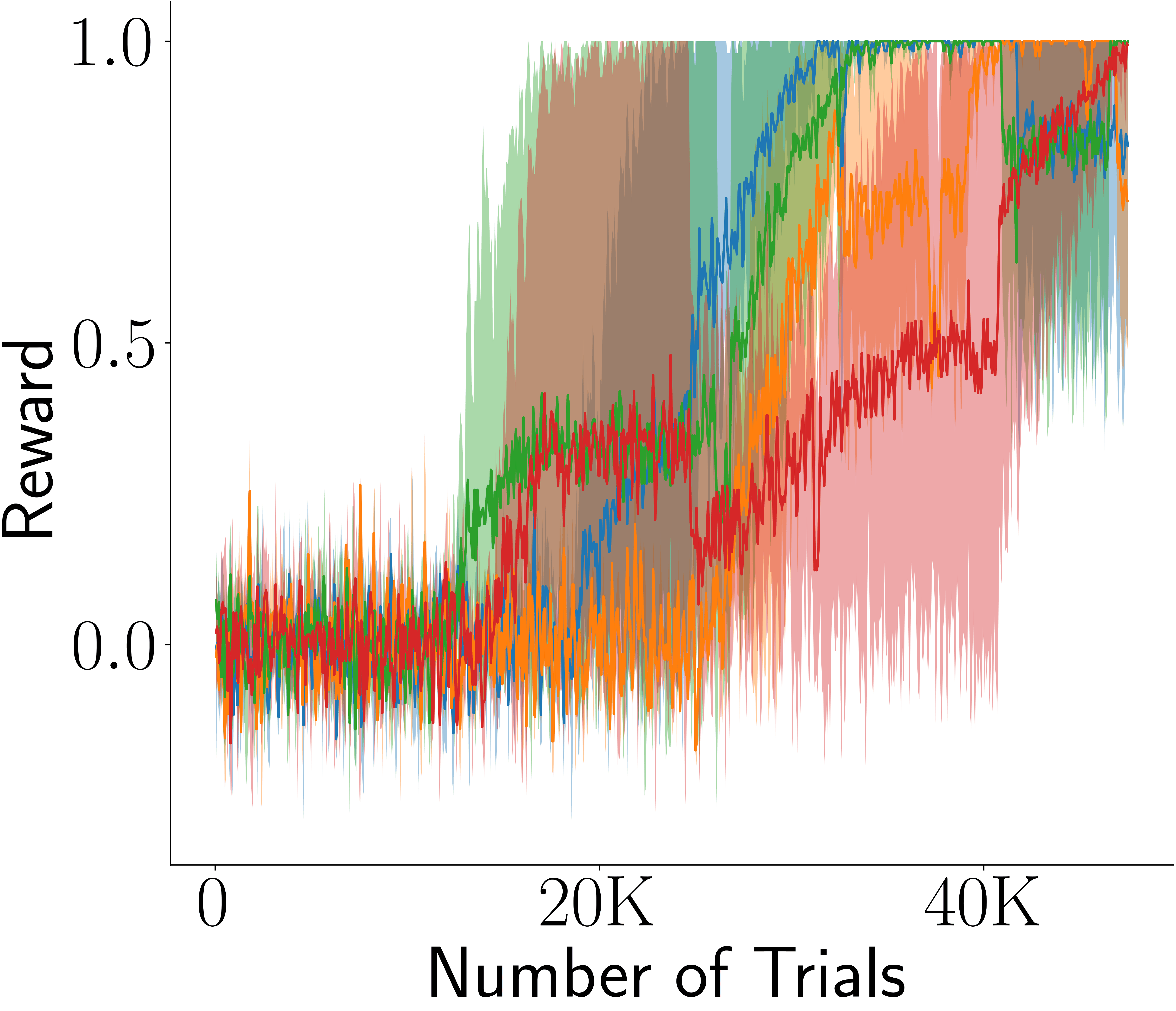} 
    \end{minipage}
    \begin{minipage}{0.23\textwidth}
        \includegraphics[width=\linewidth]{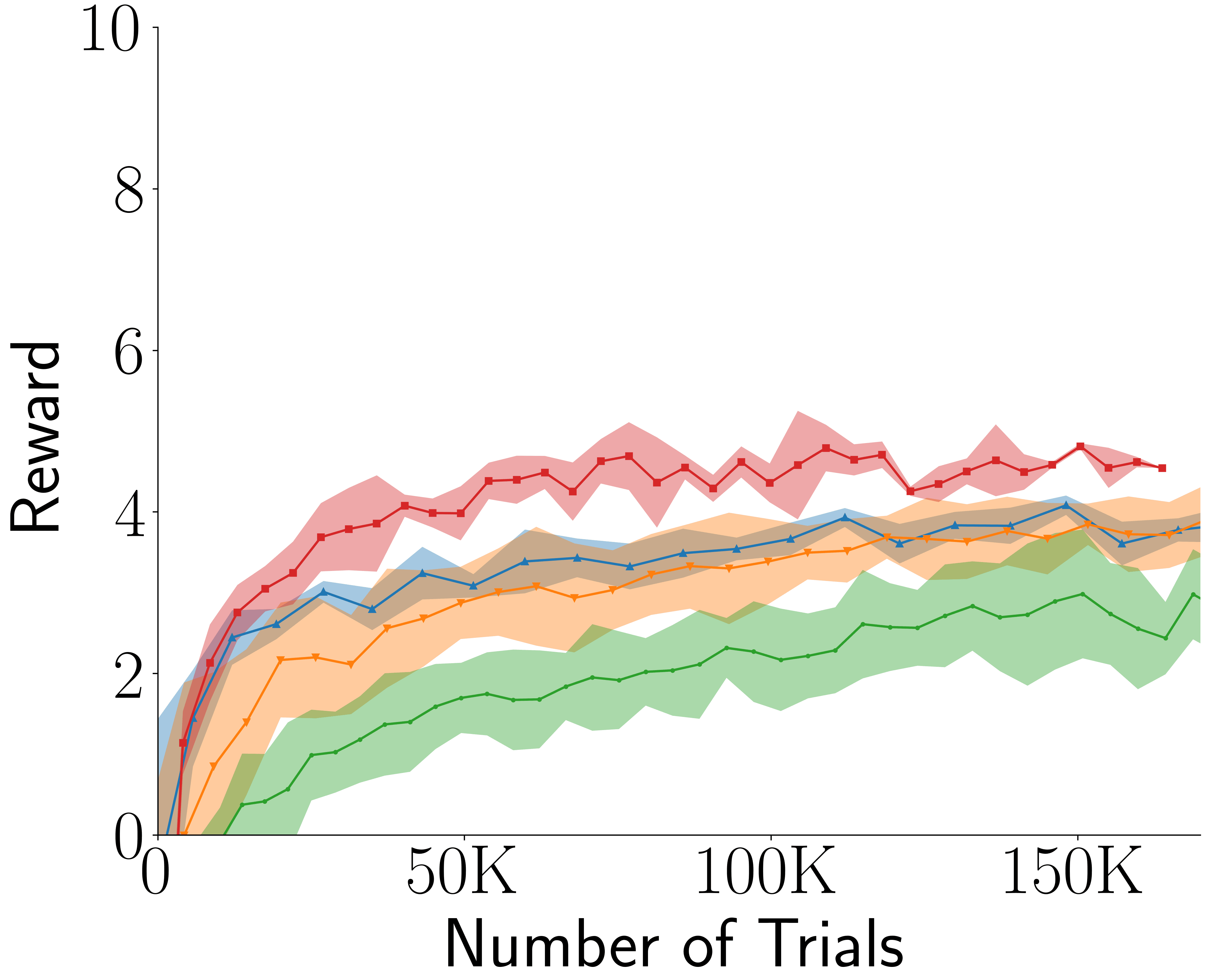} 
    \end{minipage}
    \begin{minipage}{0.4\textwidth}
        \includegraphics[width=\linewidth]{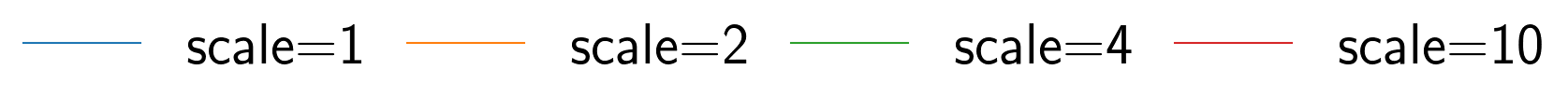} 
    \end{minipage}
    \caption{The performance (mean with standard error over five runs) across the four tasks for CogRNN and LSTM agents.}
    \label{fig:performance_results}
\end{figure*}

\section{Results}

\subsection{Agents with scale invariant memory learn equally well at different temporal scales}
We first demonstrate that when temporal relationships between task-relevant variables rescale, CogRNN agents learn at about the same rate. This is because the log compression causes the memory representation to shift rather than rescale (Fig.~\ref{fig:Cog-RNN}B), making the learning problem equally difficult at every scale where the temporal relationships fall between the hyperparameters $\taustar_{min}$ and $\taustar_{max}$.

We trained agents based on CogRNN, LSTM and RNN in each of the five environments. To evaluate robustness to rescaling in four environments (interval timing 1D and 3D, interval discrimination and delayed-match-to-sample) we conducted the training with different durations of task-relevant temporal intervals. We controlled the duration using a step size parameter, which varied from 10 to 100. In interval timing for instance, with a step size of 10, the intervals ranged from 300 to 480 steps, while with a step size of 100, they ranged from 30 to 48 steps. All other parameters of the environment and of the agents remained unchanged across different scales.

The mean reward as a function of the number of trials in those four environments is shown in Fig.~\ref{fig:performance_results} for CogRNN and LSTM agents.  
Agents based on CogRNN reached high performance in all tasks. Critically, the speed of learning was similar at different temporal scales. While LSTM and RNN agents were able to learn in all environments except the 3D interval timing environment, they did so with different learning speeds for different temporal scales.

Psychometric curves shown in Fig. S2 provide another performance measure. The y-axis on the psychometric curves represents the probability of selecting the long interval. For an agent that performs perfectly, that probability would be zero for the three short intervals and one for the three long intervals. Consistent with the results in Fig.~\ref{fig:performance_results}, CogRNN agents performed better than others. Similar to rat behavior, agents made the most mistakes for the most difficult time intervals (36 and 40 steps).

The results of training on the interval reproduction task are shown in Fig. S11, Fig. S12, Table S1, and Table S2. This task differs conceptually from the other four, since we did not train agents on rescaled versions of the environments. Instead, we trained on different interval durations, similar to \citet{deverett2019interval}. The results indicate that CogRNN agents were able to improve performance simultaneously at all training and validation intervals, demonstrating the capabilities of multi-scale learning. On the other hand, LSTM agents learned quickly only at short intervals. If LSTM agents were trained much longer, they could possibly learn the task at all scales, but critically, their learning is scale-dependent. We also highlight that training of CogRNN agents took roughly half the time as training of LSTM agents (since CogRNN has fewer trainable parameters).

\subsection{Combining scale invariant memory with translation-invariance results in invariance to temporal rescaling}
We now consider a case where the entire environment, including the duration of stimuli and the intervals between the stimuli, is rescaled in time (i.e., the elapse of time in the environment is sped up or slowed down, rather than just rescaling the time between task-relevant variables as in the previous case). Then combining scale invariant memory with convolution and maxpool results in agents which observation space is invariant (no longer covariant) of the temporal scale. If trained at one scale, such agents will follow the same policy when presented with the environment at a different scale and obtain the same performance as on the scale that was used for training (aside from edge effects mentioned in the Model section). The agents do not need to know the scale as long as the temporal relationships again fall between the hyperparameters $\taustar_{min}$ and $\taustar_{max}$. 

\begin{figure}[h!]
    \centering
    \begin{minipage}{\columnwidth}
        \centering
        \textbf{A} 
        \raisebox{-\height}{\includegraphics[width=0.9\columnwidth]{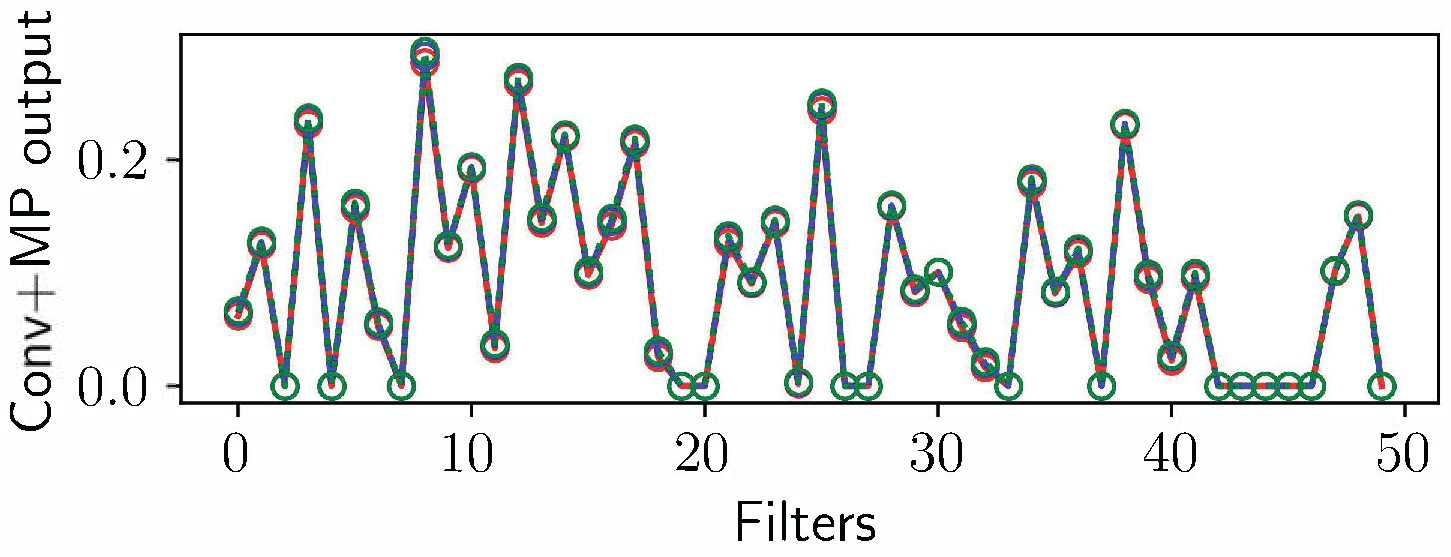}}
    \end{minipage}
    

    \begin{minipage}{\columnwidth}
        \centering
        \textbf{B} 
        \raisebox{-\height}{\includegraphics[width=0.9\columnwidth]{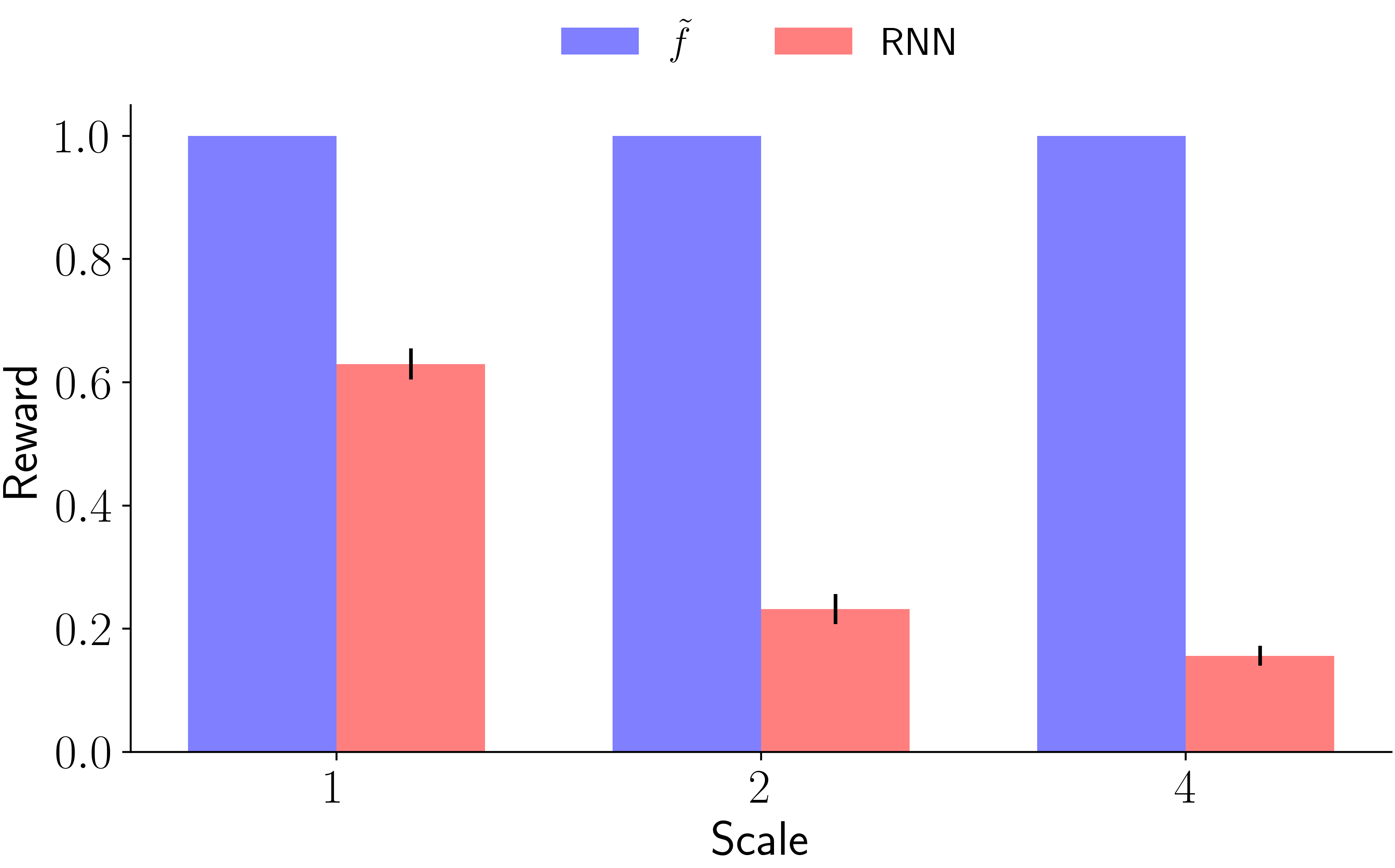}}
    \end{minipage}
    \caption{\textbf{A.} Output of convolution and pooling operations for three signals from Fig.~\ref{fig:Cog-RNN}B. \textbf{B.} Performance of CogRNN ($\tilde{f}$) and RNN agents trained on the 1D interval timing task. The agents were trained on scale 1 and evaluated on scales 1, 2 and 4.  \label{fig:results_invariance}}
\end{figure}

We demonstrate this property by training agents in the 1D interval timing environment shown in Fig.~\ref{fig:Cog-RNN}B. For simplicity, instead of A2C, we used the REINFORCE algorithm \cite{williams1992simple}. Agents with scale invariant memory followed by convolution and pooling were able to learn the task after about 100k trials. We then rescaled the time in the observation space to 2x and 4x the initial scale. Without additional training, our agents reached perfect performance (Fig.~\ref{fig:results_invariance}B). This was not the case for agents trained with RNNs, as they failed to generalize across different temporal scales. 

\subsection{Neural activity is scale invariant and resembles time cells recorded from mammalian brains}

Mammalian neural activity during tasks such as interval timing and delayed match to sample is characterized by neurons with monotonically growing/decaying firing rates and neurons that exhibit sequential activation (time cells) \cite{JinEtal09,tiganj2017sequential,tiganj2018compressed}. Consistent with the Weber-Fechner law, the width of the temporal windows in time cells increases with the peak time \citep{cao2022internally}. 

To test whether these properties from biological neurons are present in the artificial agents, we selected a representative agent from RNN, LSTM and CogRNN cores. To clean the data, for each of the three agents, we first removed neurons that were persistently active (i.e. that had constant activity, therefore not encoding any temporal information) and neurons that were completely silent. We then identified neurons that had monotonically growing/decaying activity, finding such neurons in all three agents (Fig. S3). 

\begin{figure*}[h!]
    \centering
        \begin{tabular}{c c c}
             \ \ \ \ RNN &  \ \ \ \ \  LSTM & \ \ \ \ \ \ CogRNN \\ 
            \includegraphics[scale=0.32]{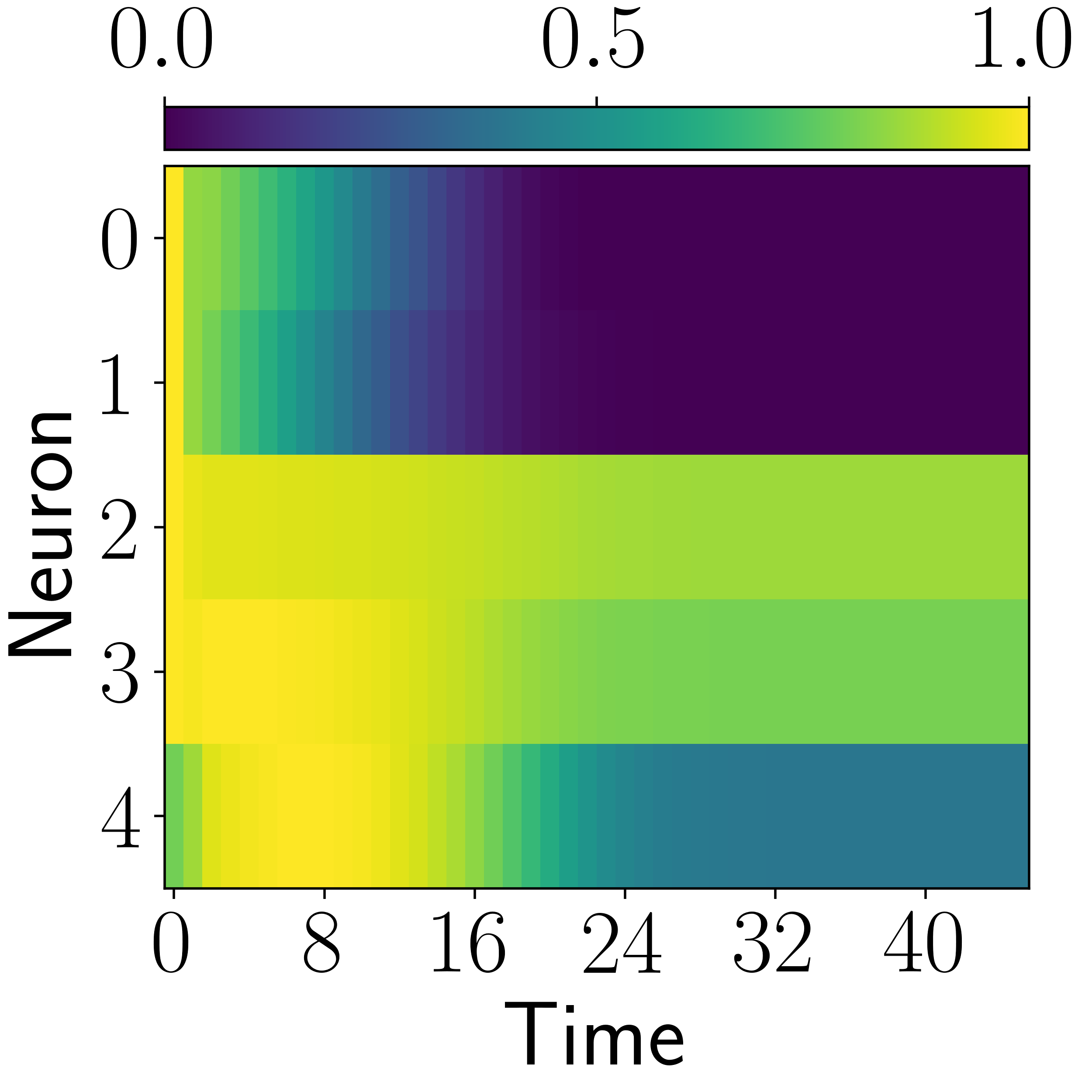} &    
            \includegraphics[scale=0.32]{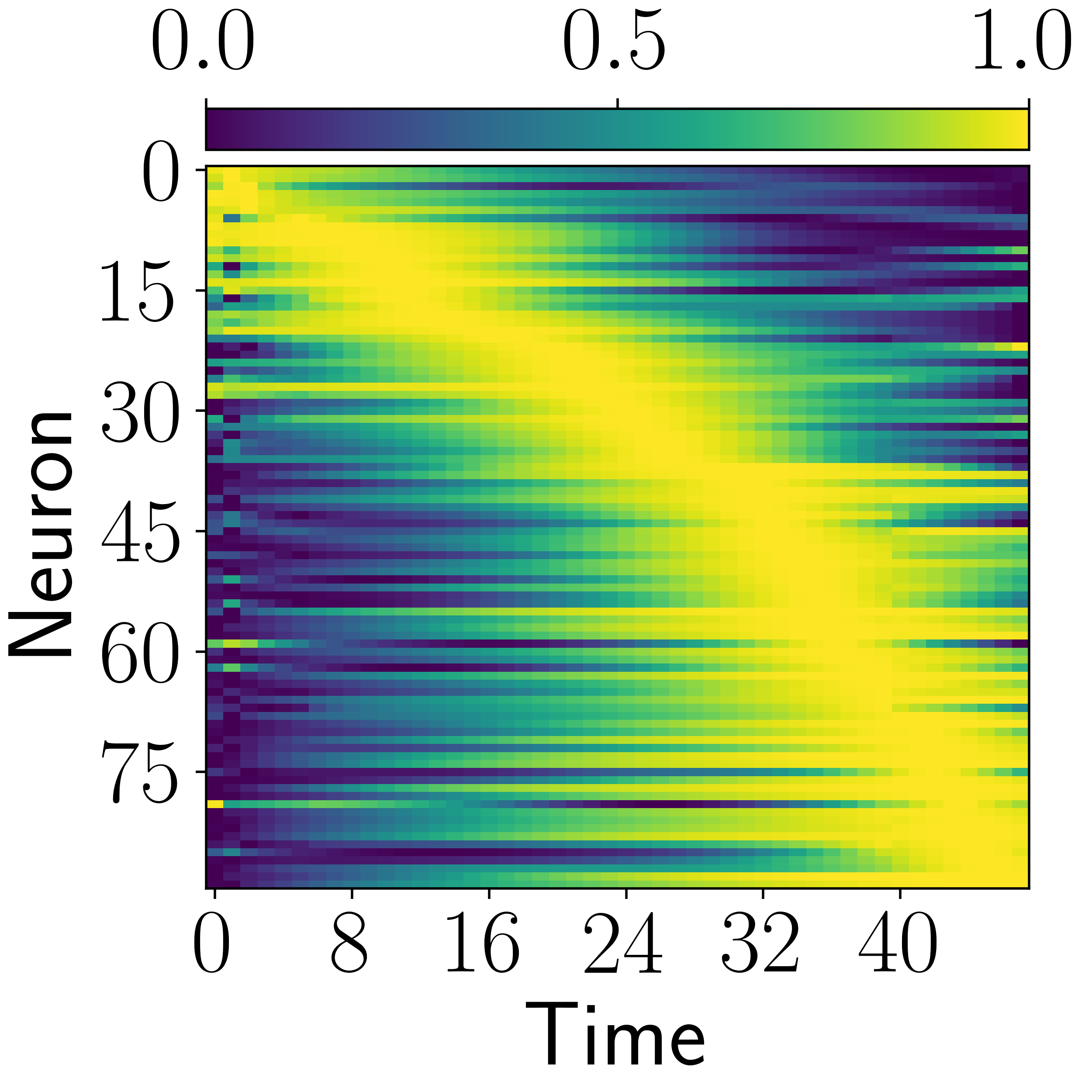} & 
            \includegraphics[scale=0.32]{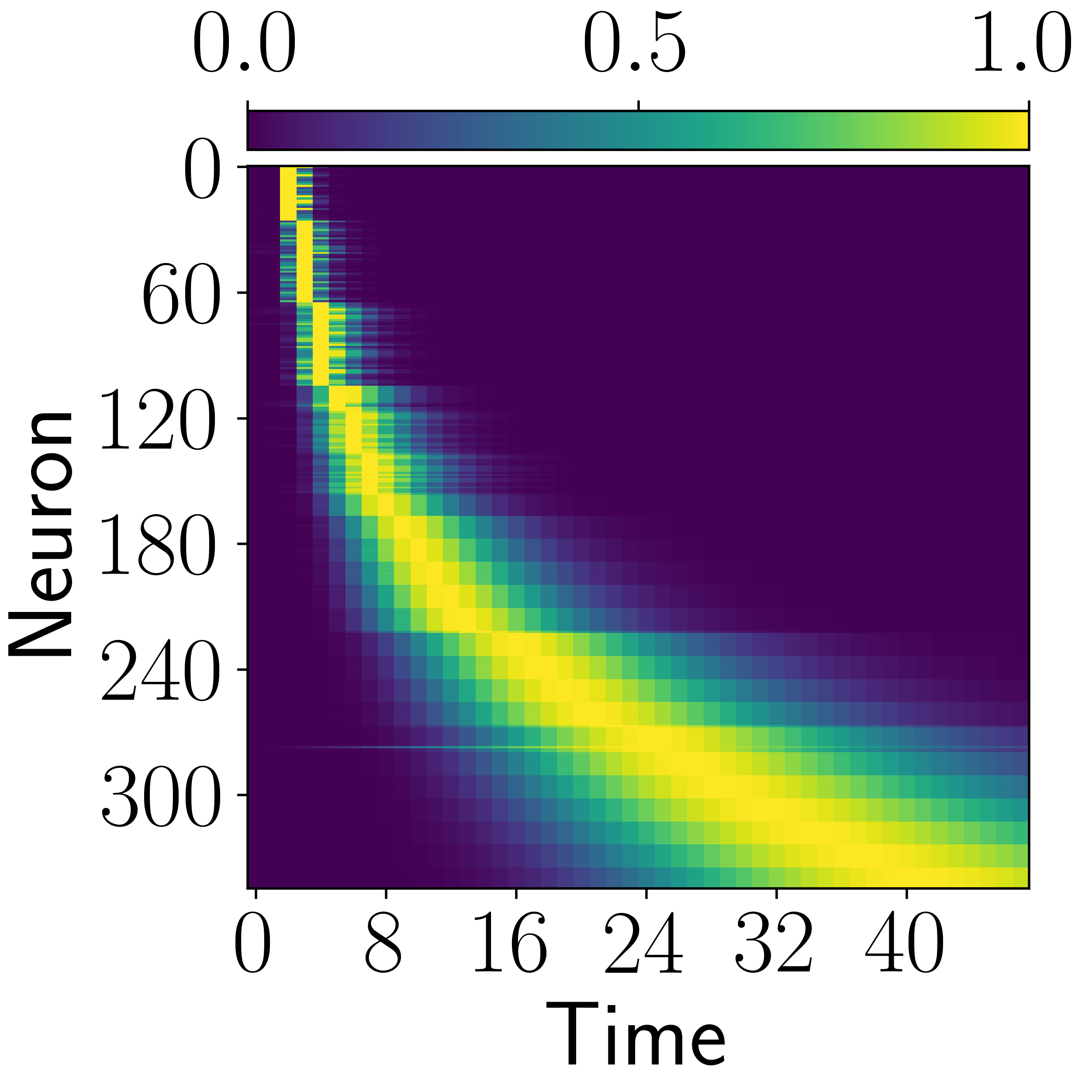}\\ 
        \end{tabular}
    \caption{Normalized activity of neurons that resemble time cells from three representative agents. Neurons are sorted by peak time.}
    \label{fig:neurons_timecells}
\end{figure*}

Next, we identified neurons with transient activation sometime during the timed interval (neurons that resemble time cells) and again found them in all three agents (Fig.~\ref{fig:neurons_timecells}). For CogRNN, the activation pattern resembles the scale invariant impulse responses shown in Fig.~\ref{fig:Cog-RNN}A. 

To better understand the activation profiles, we fitted each time cell with a Gaussian distribution and estimated standard deviation. For a representation to be scale invariant, the relationship between the peak time and the standard deviation should be linear. This was the case only for CogRNN neurons (Fig. S4). Note that some of the CogRNN neurons with large peaks deviate from the linear relationship. This is because the numerical derivative $d^k/ds^k$ suffers from edge effects. Increasing $\taustar_{max}$ would extend the range of CogRNN representation while requiring a logarithmic increase in the number of neurons.

\subsection{Ablation studies}
To validate the robustness of our approach and understand better potential limitations,  we ran several ablation studies that are added to the Supplemental Information. The ablations studies were conducted on the 3D interval timing environment. We trained the agents using LSTM and RNN with frozen weights, neither of which was able to learn the task (Fig. S1). Note that CogRNN can be considered a type of RNN with frozen (analytically calculated) weights. 

We removed the inverse Laplace transform and instead of $\tilde{f}$ used CogRNN with $F$ (the Laplace transform). CogRNN $F$ agents were also not able to learn the task as well as $\tilde{f}$ (Fig. S1), indicating the importance of the inverse Laplace transform and the time cells (since those are generated through the inverse Laplace transform). 

To examine the impact of CogRNN parameters we changed the values of $\taustar_{min}$ (Fig. S5), $\taustar_{max}$ (Fig. S6), and $k$ (Fig. S7). In general, CogRNN $\tilde{f}$ agents reached high performance except in cases where hyperparameters were inappropriate for the properties of the environment.

We further evaluated the robustness of the results to the choice of the training parameters by varying learning rate (Fig. S8), entropy (Fig. S9) and horizon (Fig. S10). In general, for our main experiments, we selected the hyperparameters that led to the best results as specified in the Model section.

\section{Discussion}
Our results demonstrated that artificial agents can approach scale invariant learning when equipped with scale invariant memory. When temporal relationships in the environment rescale, the log compression in the scale invariant memory causes temporal memory to shift rather than rescale. This shift ensures the difficulty of learning the task remains consistent. We have demonstrated that when the environment is perfectly rescaled, the agents can generalize and use the knowledge acquired at one scale to solve the problem at different scales. These results were observed despite the fact that RL algorithms such as A2C contain components that break scale invariance, including exponential temporal discounting, horizon (n-step returns) and rollout length. Combining the proposed approach with scale invariant (power-law) temporal discounting \cite{TanoEtal20,tiganj2019estimating,redish2010neural} could further improve the results.

We believe that the fundamental principles behind this work have the potential to impact research above and beyond interval timing, temporal discrimination and delayed match to sample. Humans can move objects from one place on the desk to another, furniture from one room to another, or drive a car from one location to another. We did not evolve to drive cars for hundreds of miles. But an ample amount of data from cognitive psychology suggests that we did evolve a scale invariant representation that enables us to efficiently and automatically represent arbitrary scales and operate on them. Building scale invariance into deep neural networks is critical for developing systems that can adjust to new environments rapidly and flexibly without the need for continual tweaking of hyperparameters. 

\section{Acknowledgments}
This research was supported in part by Lilly Endowment, Inc., through its support for the Indiana University Pervasive Technology Institute.

\bibliography{references,bibdesk}


\newpage
\appendix
\onecolumn

\setcounter{page}{1} 
\renewcommand{\thepage}{S\arabic{page}}

\section{Supplemental Information}
\vspace{0.7cm}

\setcounter{figure}{0}
\renewcommand{\thefigure}{S\arabic{figure}}

\setcounter{table}{0}
\renewcommand{\thetable}{S\arabic{table}}

\begin{figure*}[h!]
    \centering
    \includegraphics[width=0.4\textwidth]{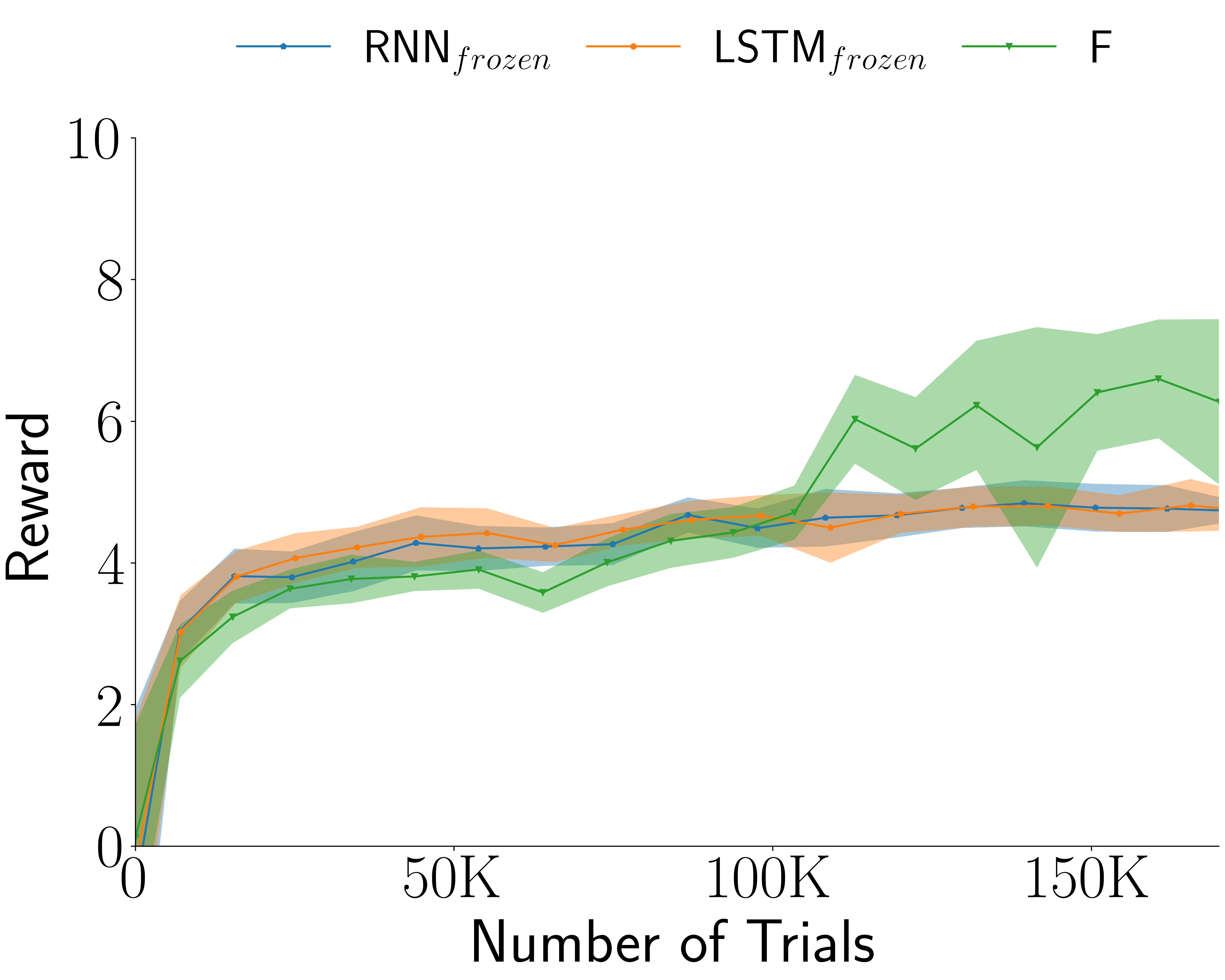}
    \caption{Performance of the frozen RNN, frozen LSTM, and CogRNN F agents for the same time scale during the 3D interval timing task.}
    \label{fig:performance_ablation}
\end{figure*}

\begin{figure}[h!]
    \centering
        \begin{tabular}{c c c}
             RNN & LSTM & CogRNN \\ 
            \includegraphics[scale=0.32]{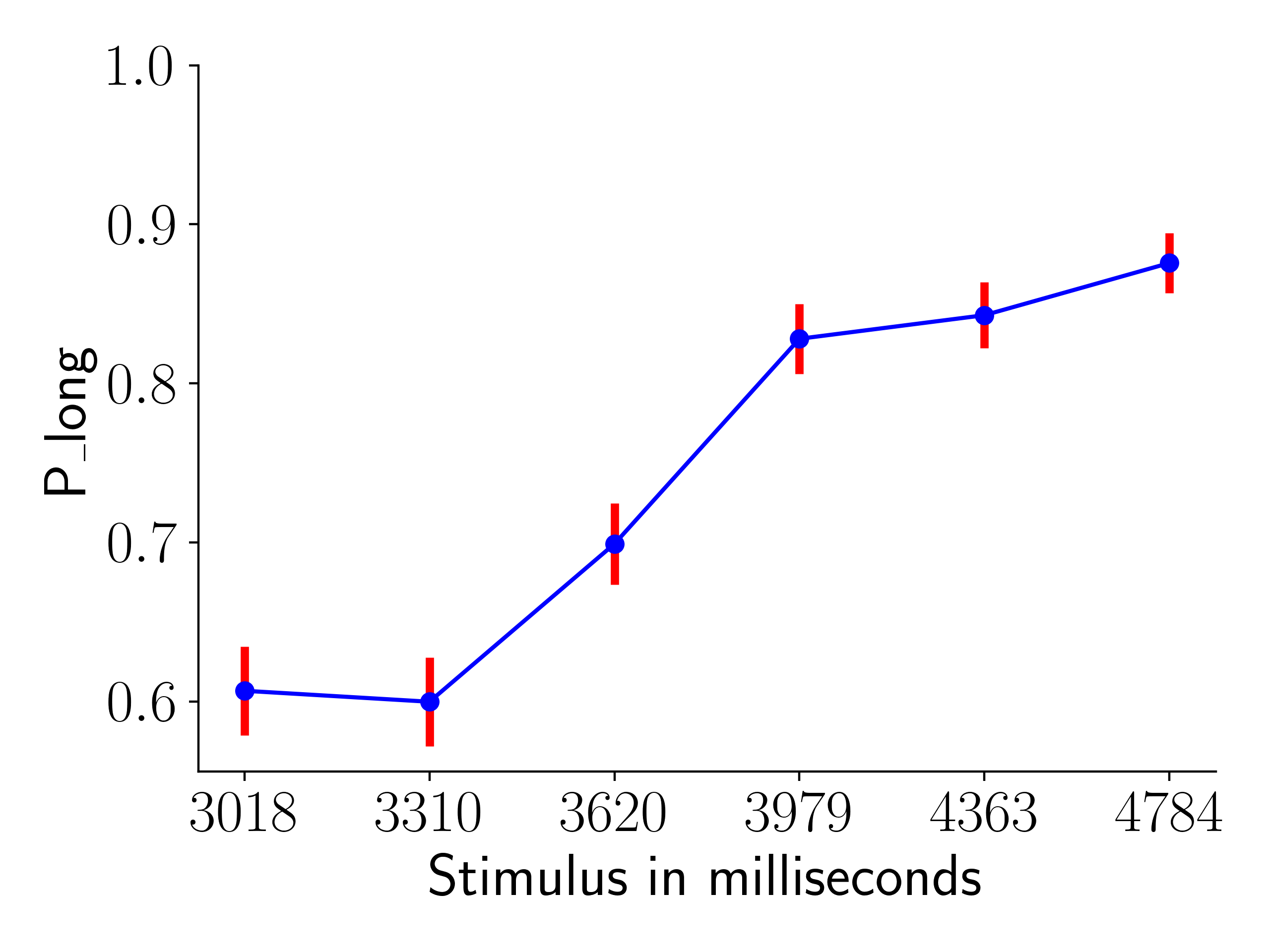} &    
            \includegraphics[scale=0.32]{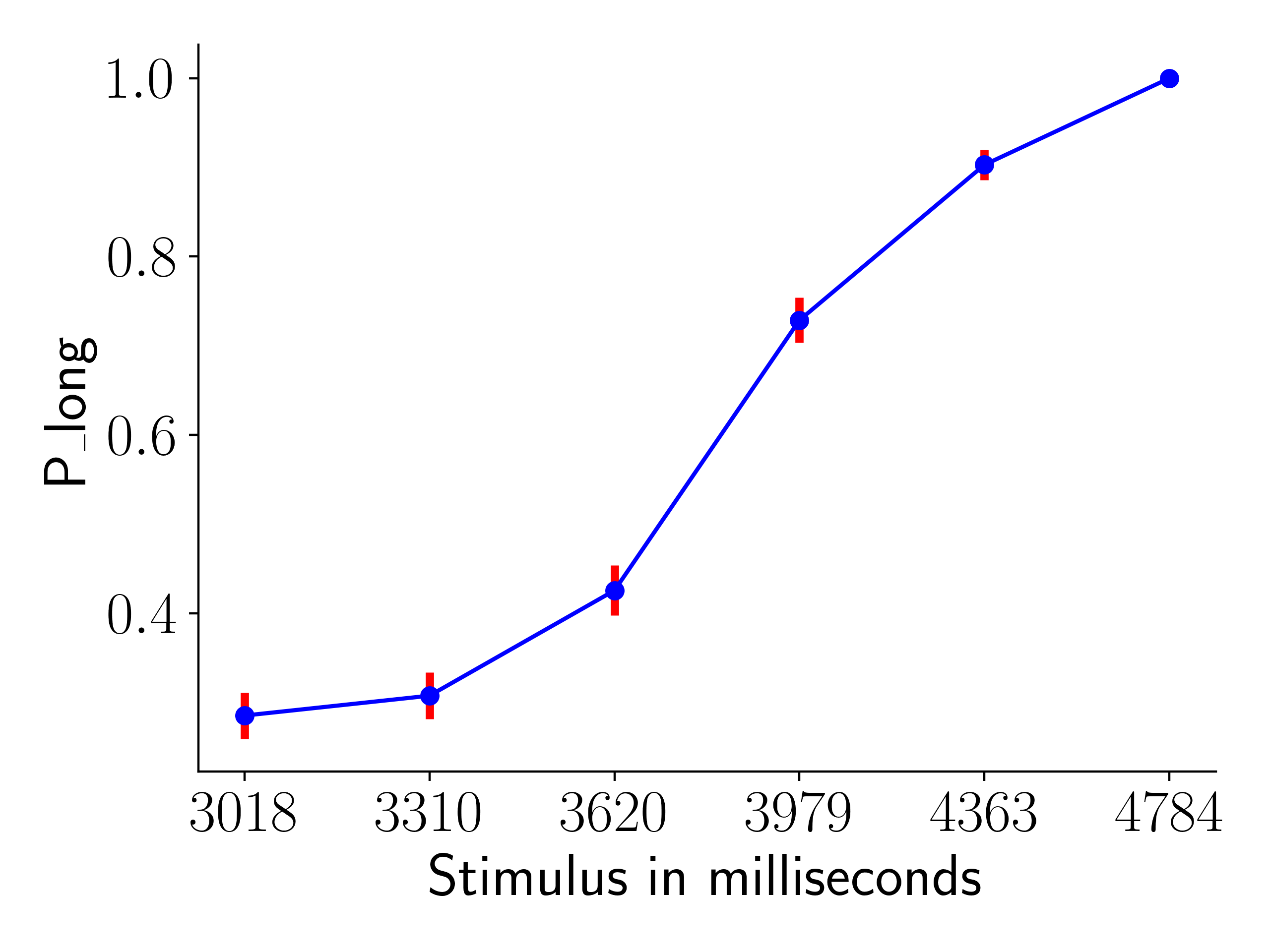} & 
            \includegraphics[scale=0.32]{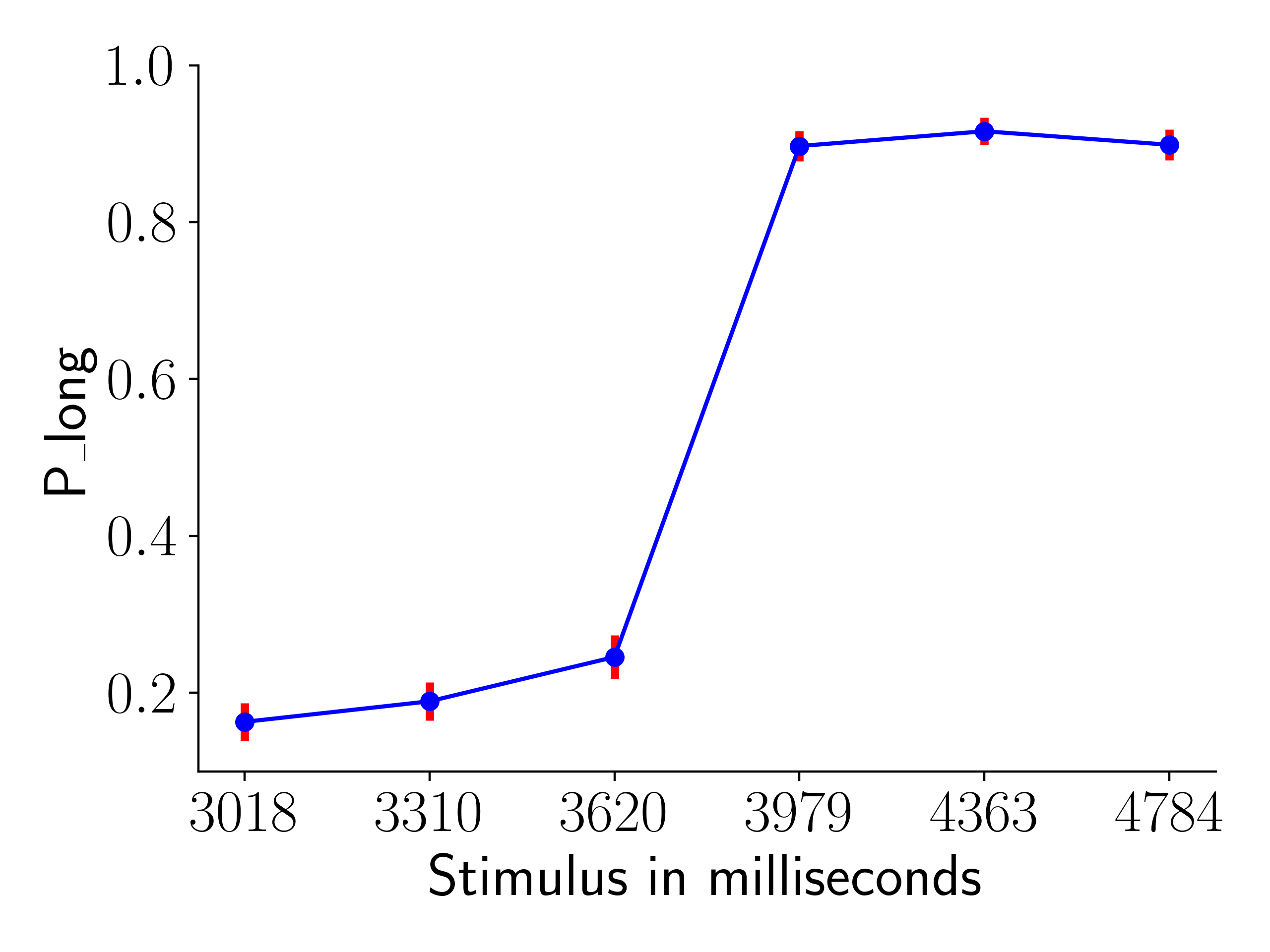}\\ 
        \end{tabular}
    \caption{Psychometric curves for (a) RNN, (b) LSTM, and (c) CogRNN $\tilde{f}$ agents during 3D interval timing task. The average performance across agents is shown for each group, with error bars indicating the confidence intervals.}
        \label{fig:psychometric_supplemental}
\end{figure}

\begin{figure}[h!]
    \centering
        \begin{tabular}{c c c}
             \ \ \ \ \ \ RNN & \ \  \ \ \ \ \  LSTM & \ \ \ \ \ \ \ \ CogRNN\\ 
            \includegraphics[scale=0.32]{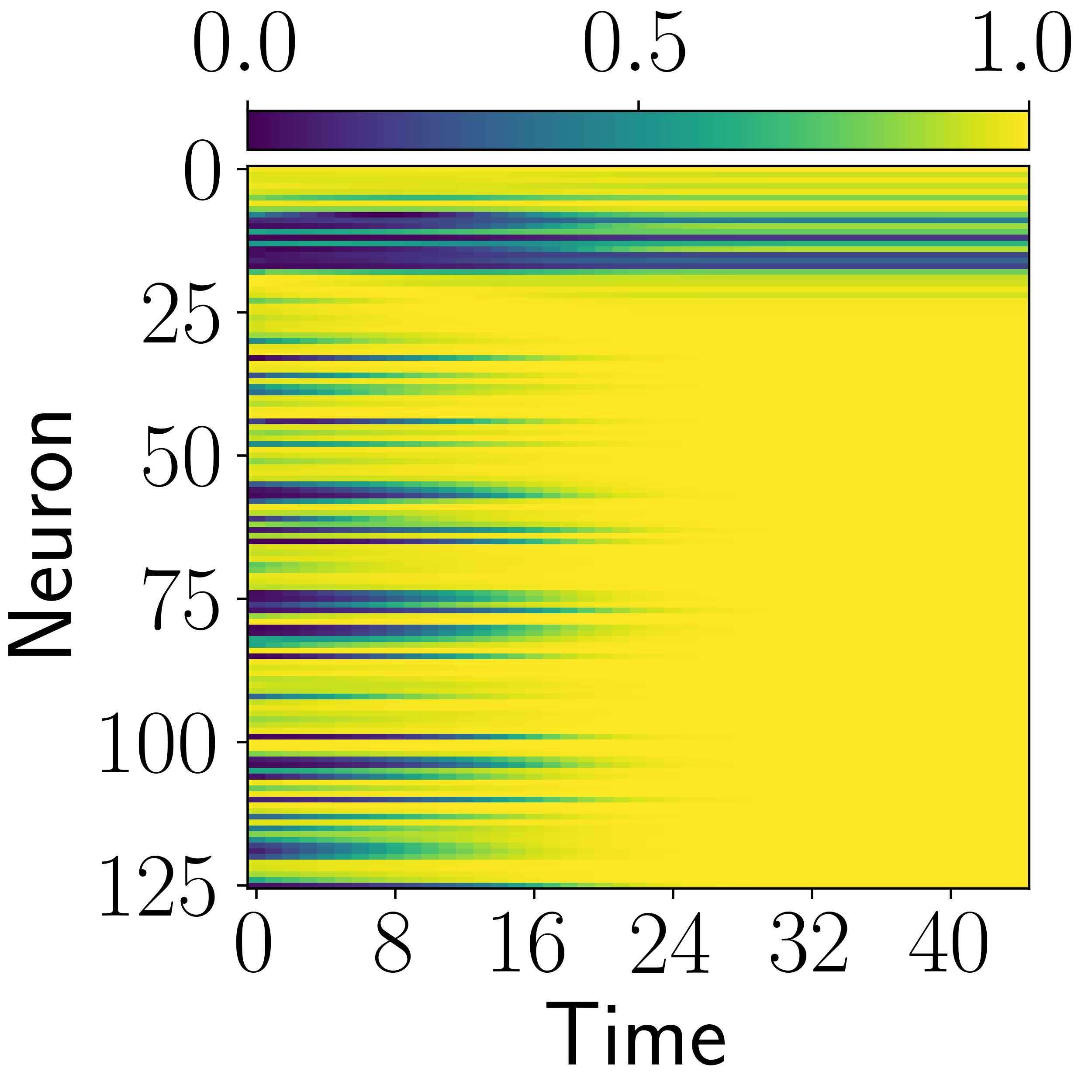} &    
            \includegraphics[scale=0.32]{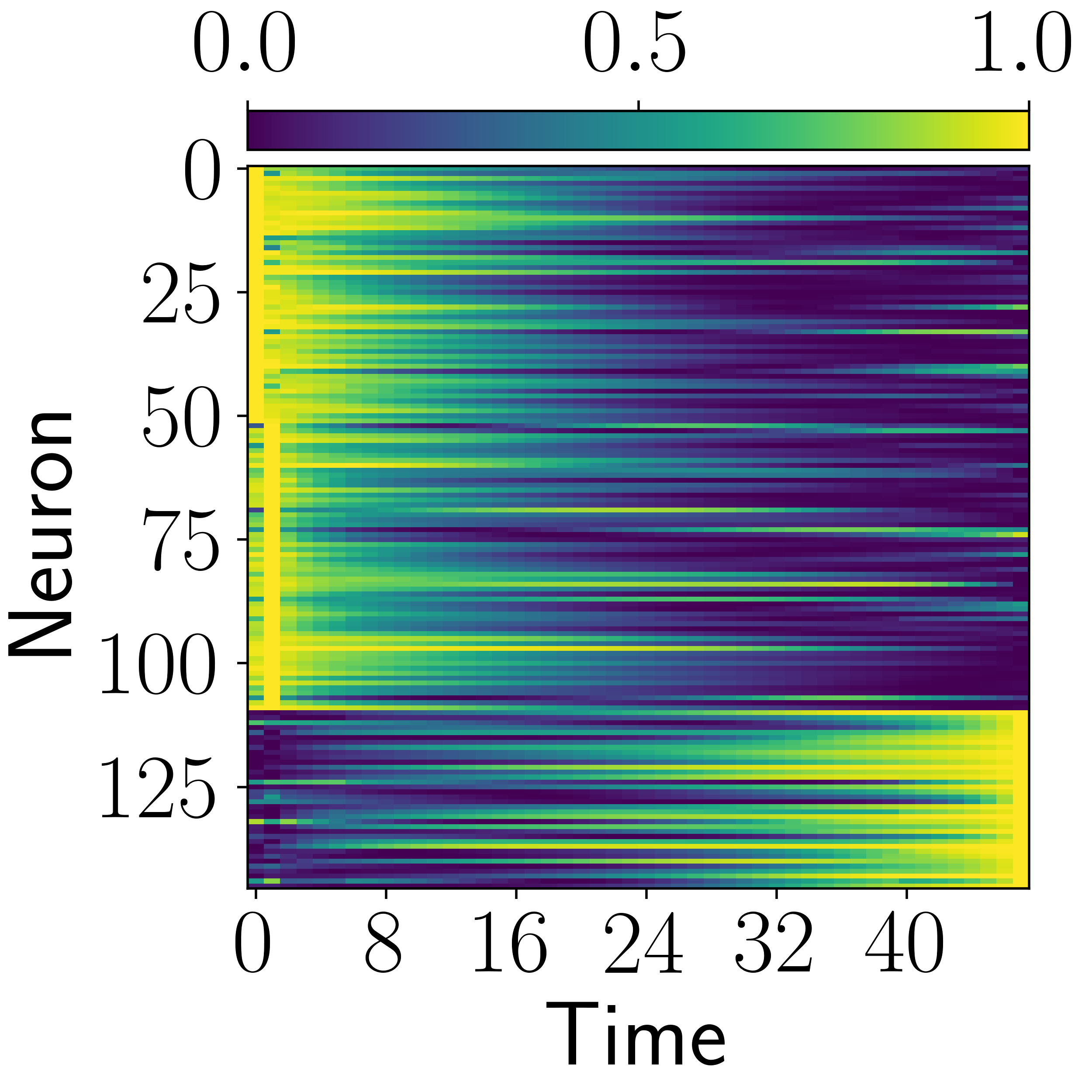} & \includegraphics[scale=0.32]{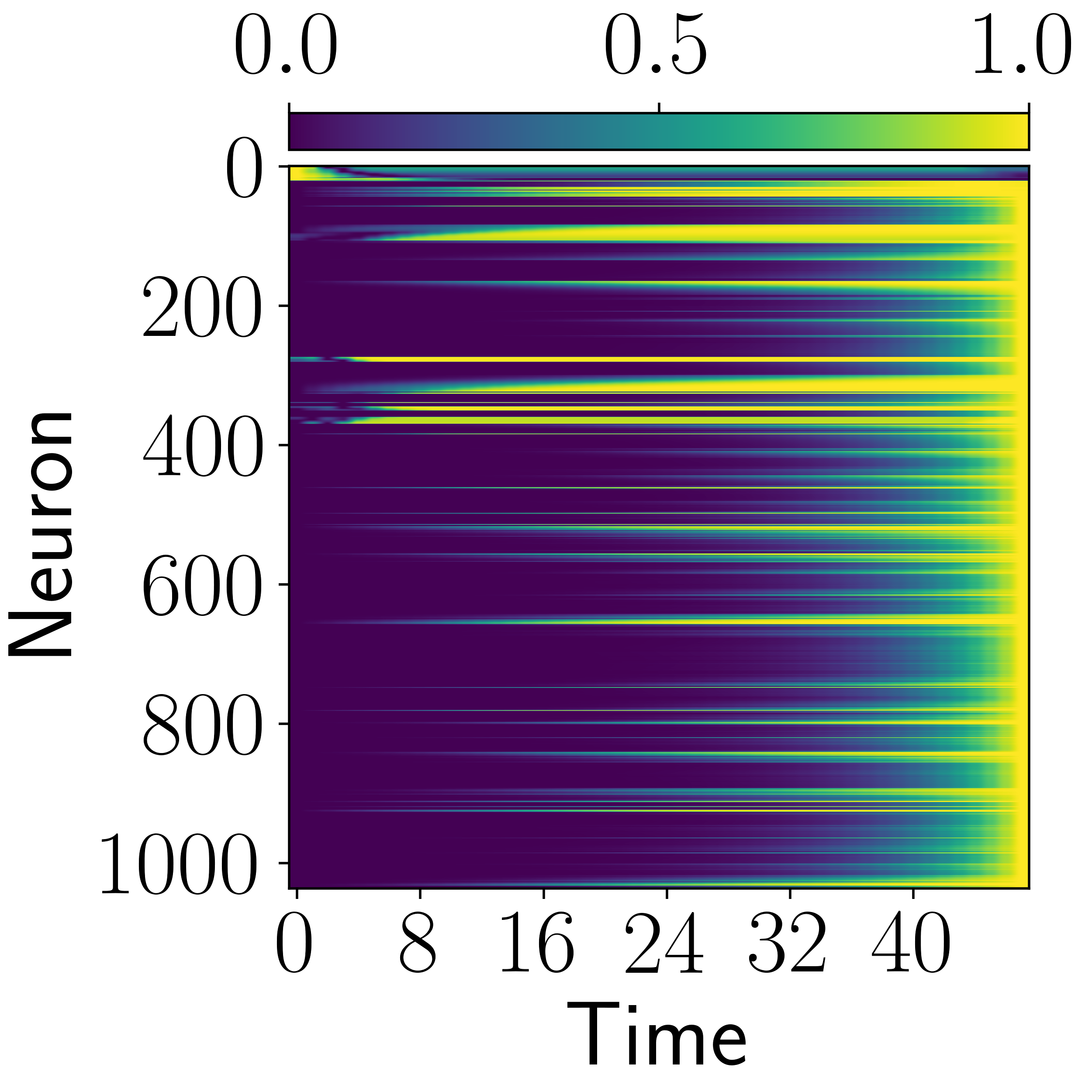}\\ 
        \end{tabular}
    \caption{Normalized activity of neurons that have monotonically decreasing/increasing activity from three representative agents. Neurons are sorted by peak time.}
    \label{fig:neurons_decay_ramp}
\end{figure}

\begin{figure}[h!]
    \centering
        \begin{tabular}{c c c}
             \hspace{22pt} RNN & \hspace{22pt} LSTM & \hspace{22pt} CogRNN \\ 
            \includegraphics[scale=0.32]{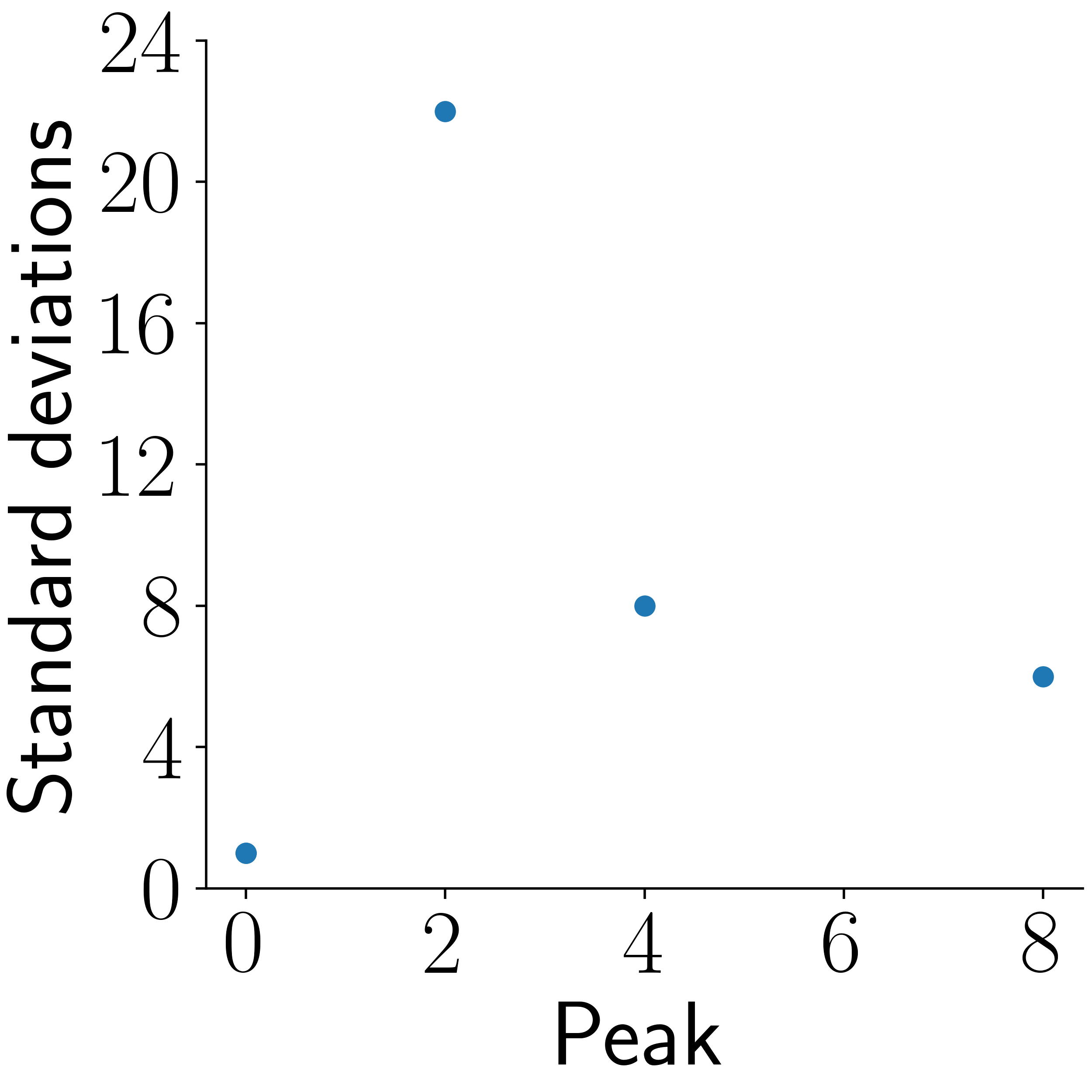} &    
            \includegraphics[scale=0.32]{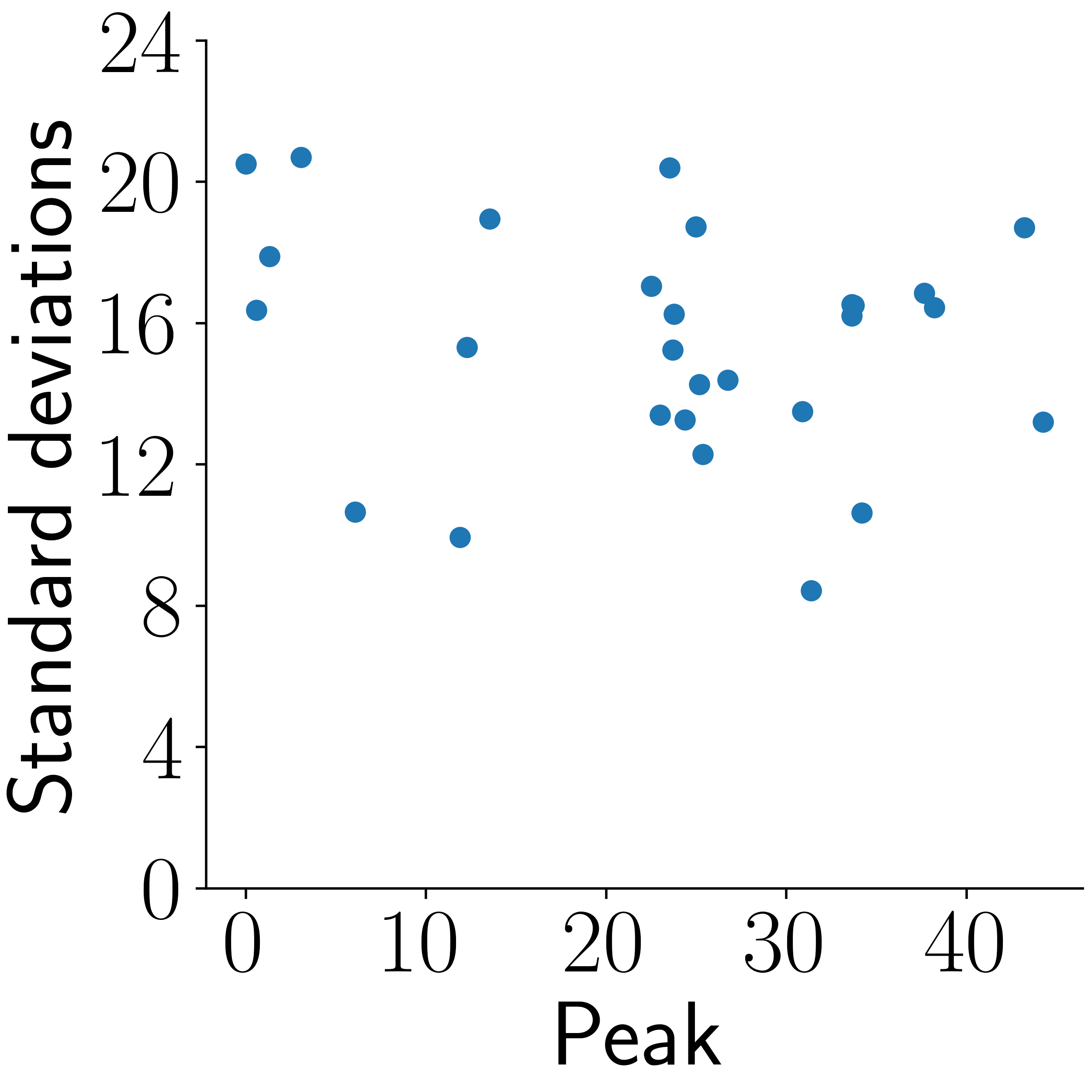} & \includegraphics[scale=0.32]{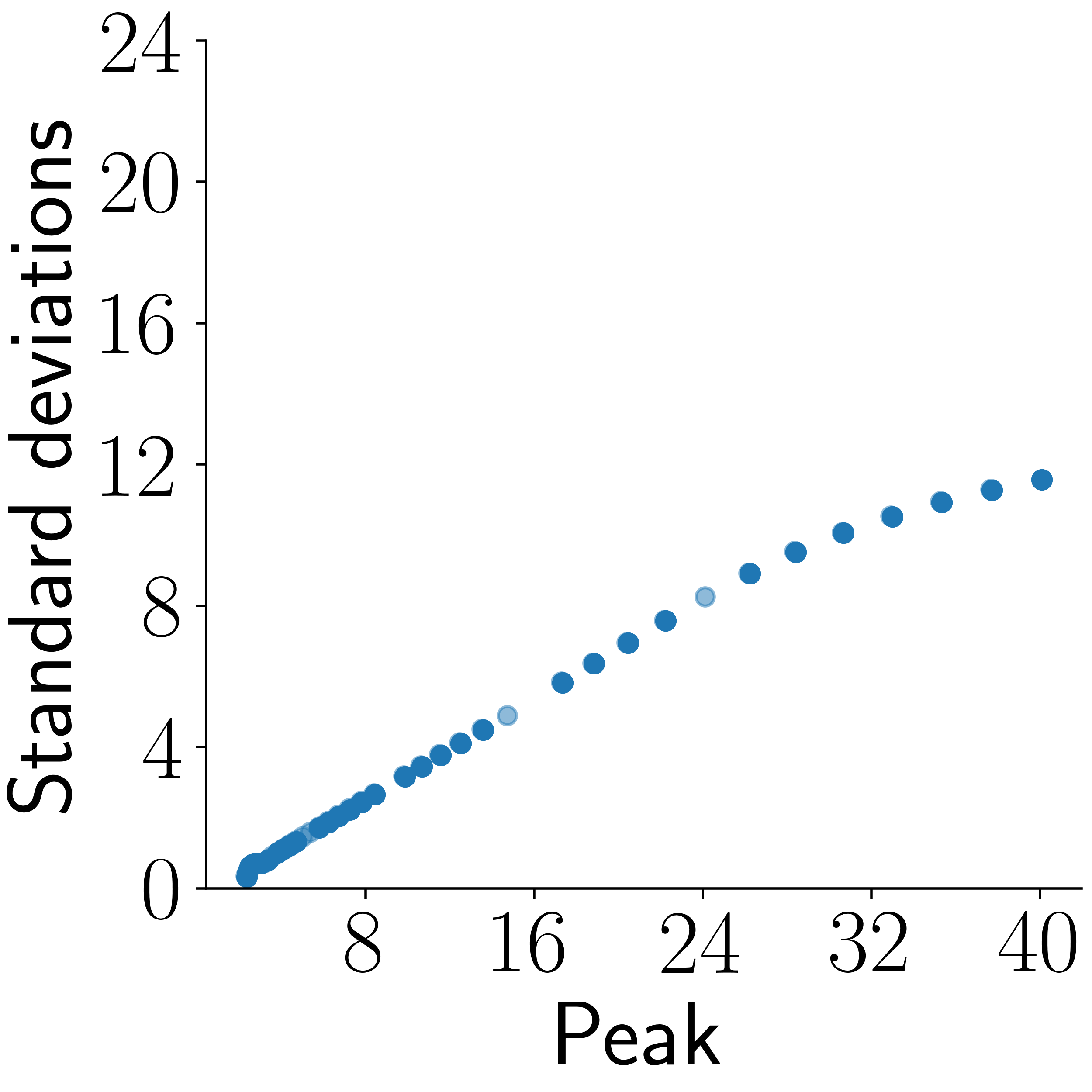}\\ 
        \end{tabular}
    \caption{Relationship between peak time and standard deviation for neurons whose activity resembled time cells.}
    \label{fig:peaks_width}
\end{figure}

\begin{figure*}[h!]
    \centering
    \includegraphics[width=0.4\textwidth]{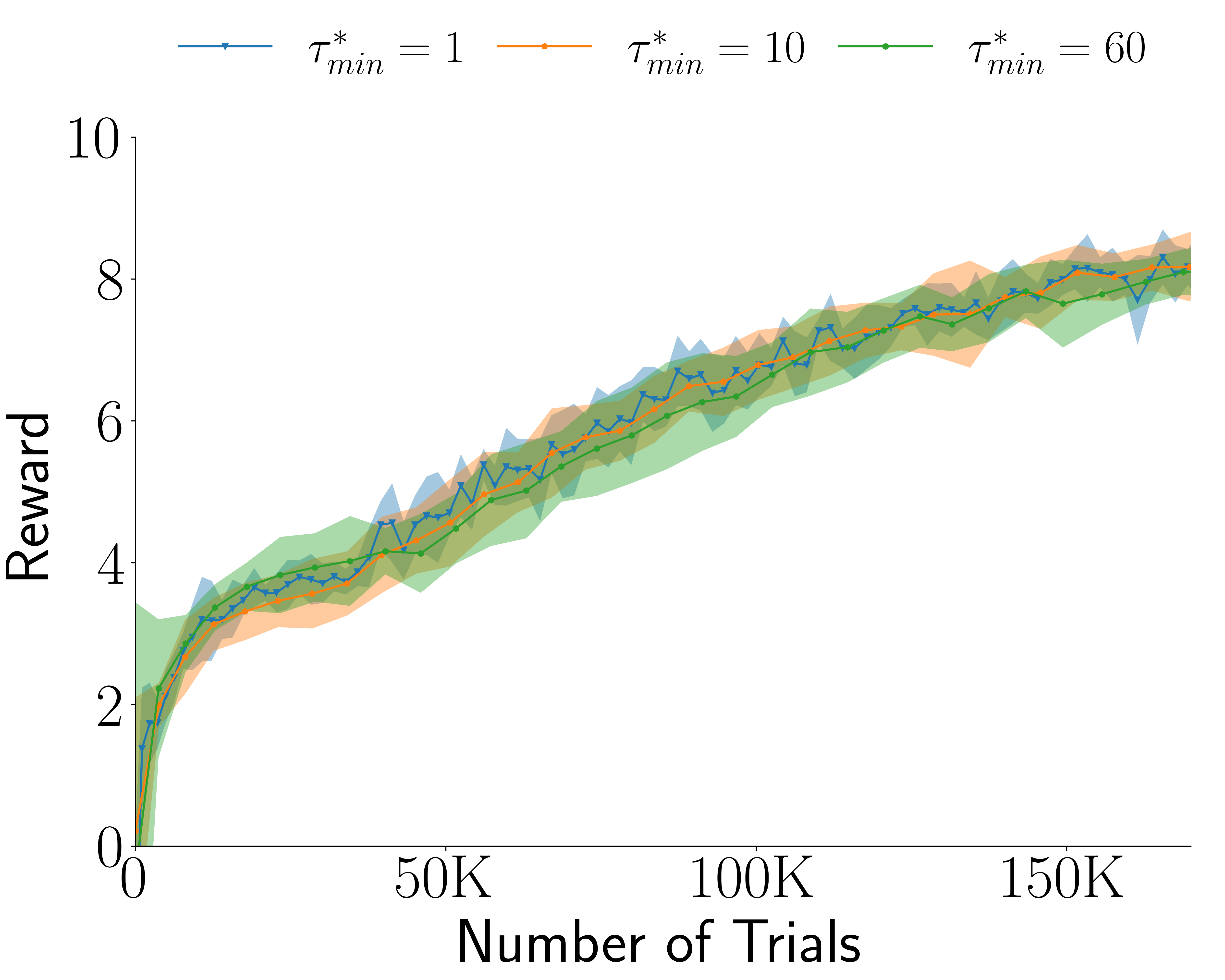}
    \caption{Performances of the CogRNN $\tilde{f}$ agents for different values of $\taustar_{min}$.}
    \label{fig:ablation_f_tilde_tstr_min}
\end{figure*}

\begin{figure*}[h!]
    \centering
    \includegraphics[width=0.4\textwidth]{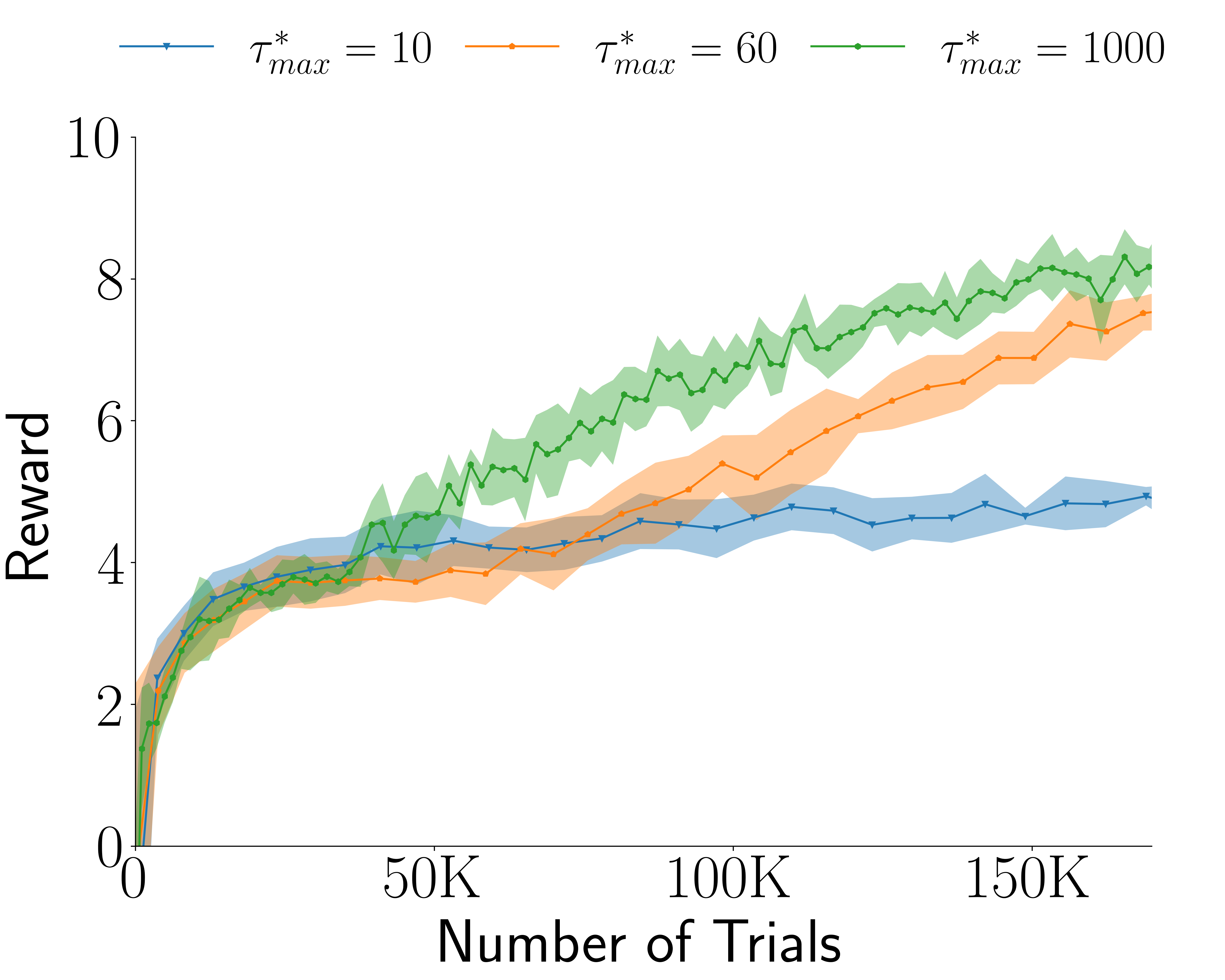}
    \caption{Performances of the CogRNN $\tilde{f}$ agents for different values of $\taustar_{max}$.}
    \label{fig:ablation_f_tilde_tstr_max}
\end{figure*}

\begin{figure*}[h!]
    \centering
    \includegraphics[width=0.4\textwidth]{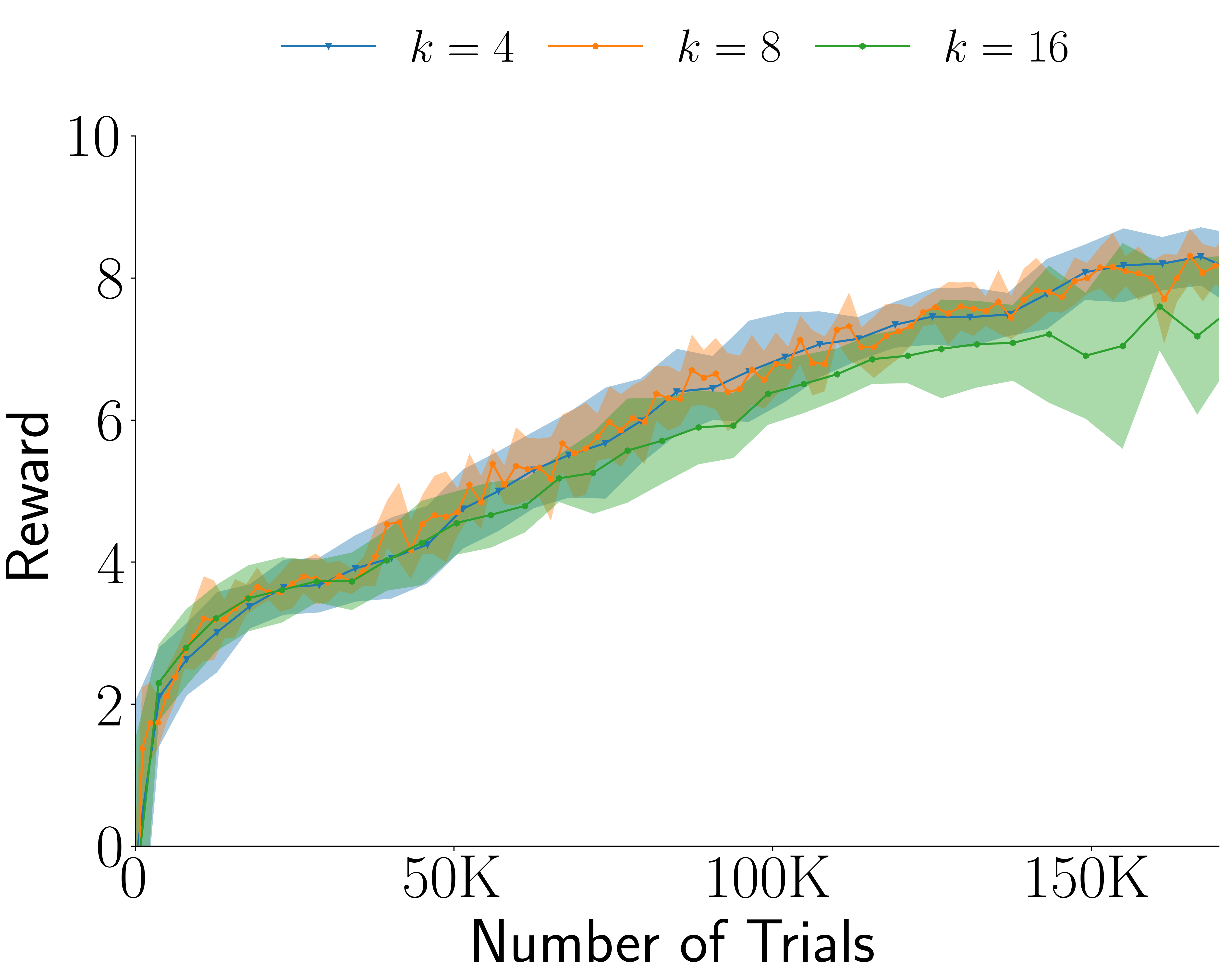}
    \caption{Performances of the CogRNN $\tilde{f}$ agents for different values of $k$.}
    \label{fig:ablation_f_tilde_k}
\end{figure*}

\begin{figure*}[h!] 
    \centering
    \begin{tabular}{c}  
        \includegraphics[scale=0.23]{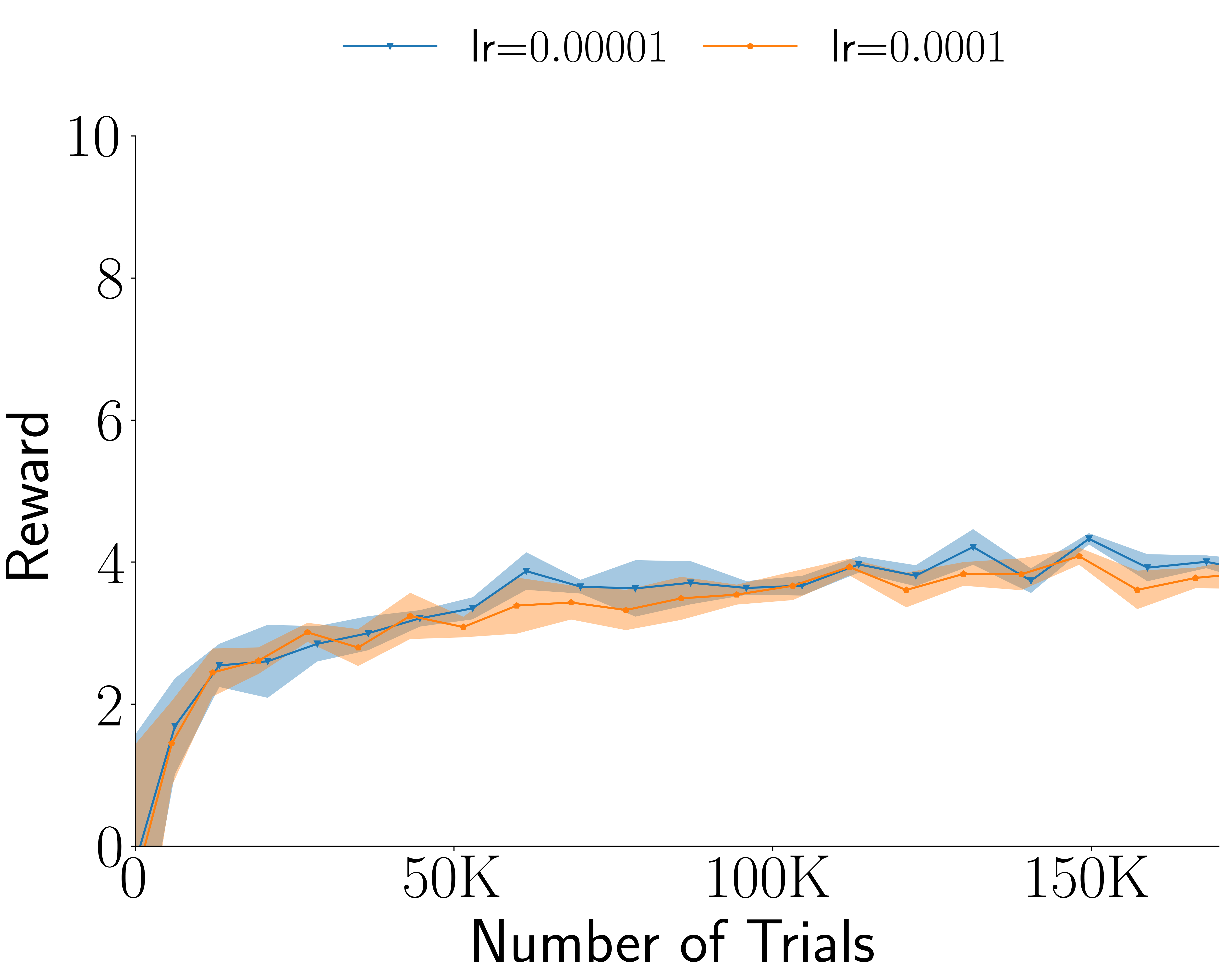}  \\
        \LARGE{a}\\
        \includegraphics[scale=0.23]{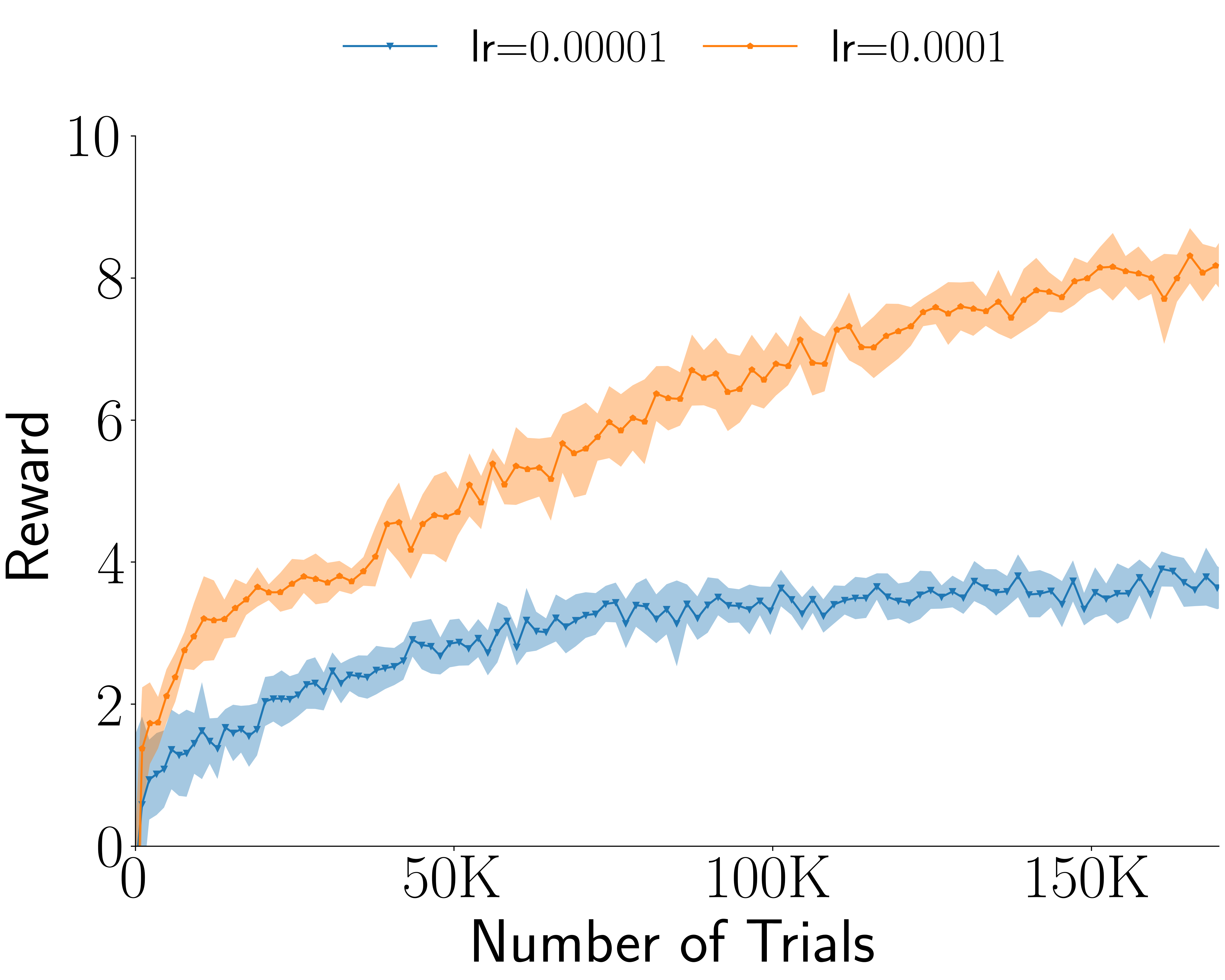} \\
        \LARGE{b}\\

        \end{tabular} 
        \caption{Performances of (a) LSTM and (b) CogRNN $\tilde{f}$ agents for different learning rates.}
        \label{fig:learning_rate_supplemental}
\end{figure*}

\begin{figure*}[h!] 
    \centering
    \begin{tabular}{c}  
        \includegraphics[scale=0.23]{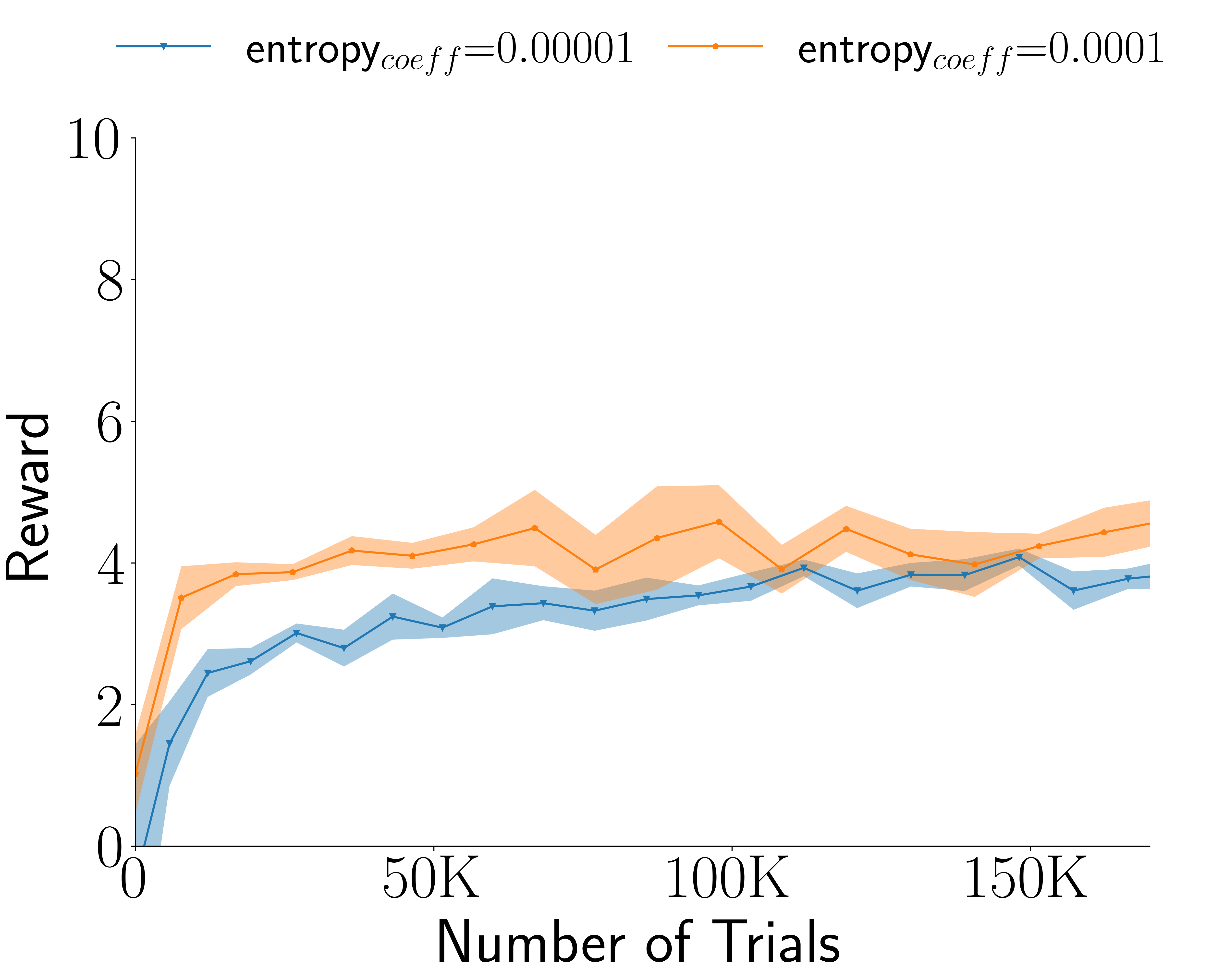}  \\
        \LARGE{a}\\
        \includegraphics[scale=0.23]{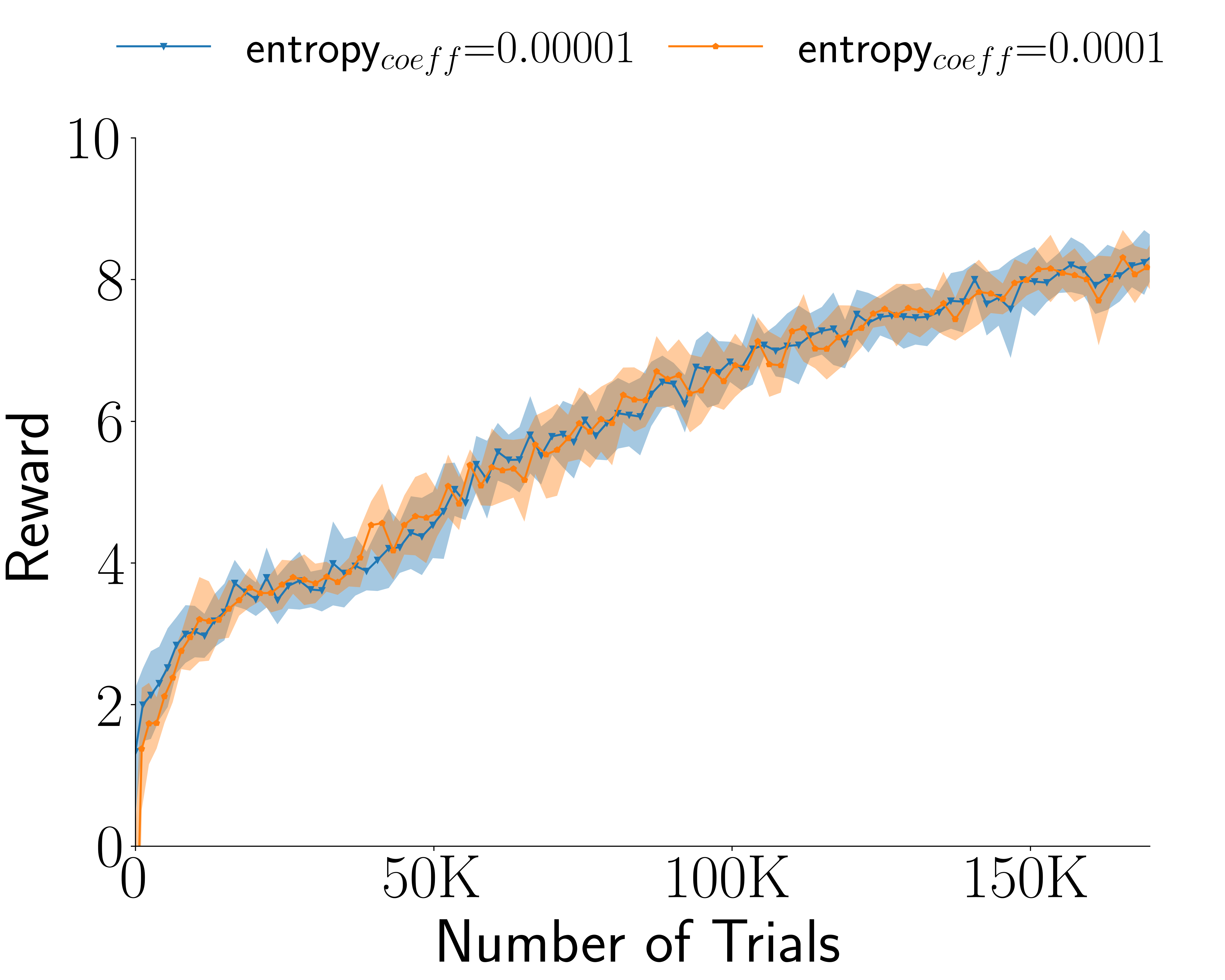} \\
        \LARGE{b}\\
        \end{tabular} 
        \caption{Performances of (a) LSTM and (b) CogRNN $\tilde{f}$ agents for different entropy coefficients.}
        \label{fig:entropy_supplemental}
\end{figure*}

\begin{figure*}[h!] 
    \centering
    \begin{tabular}{c}  
        \includegraphics[scale=0.23]{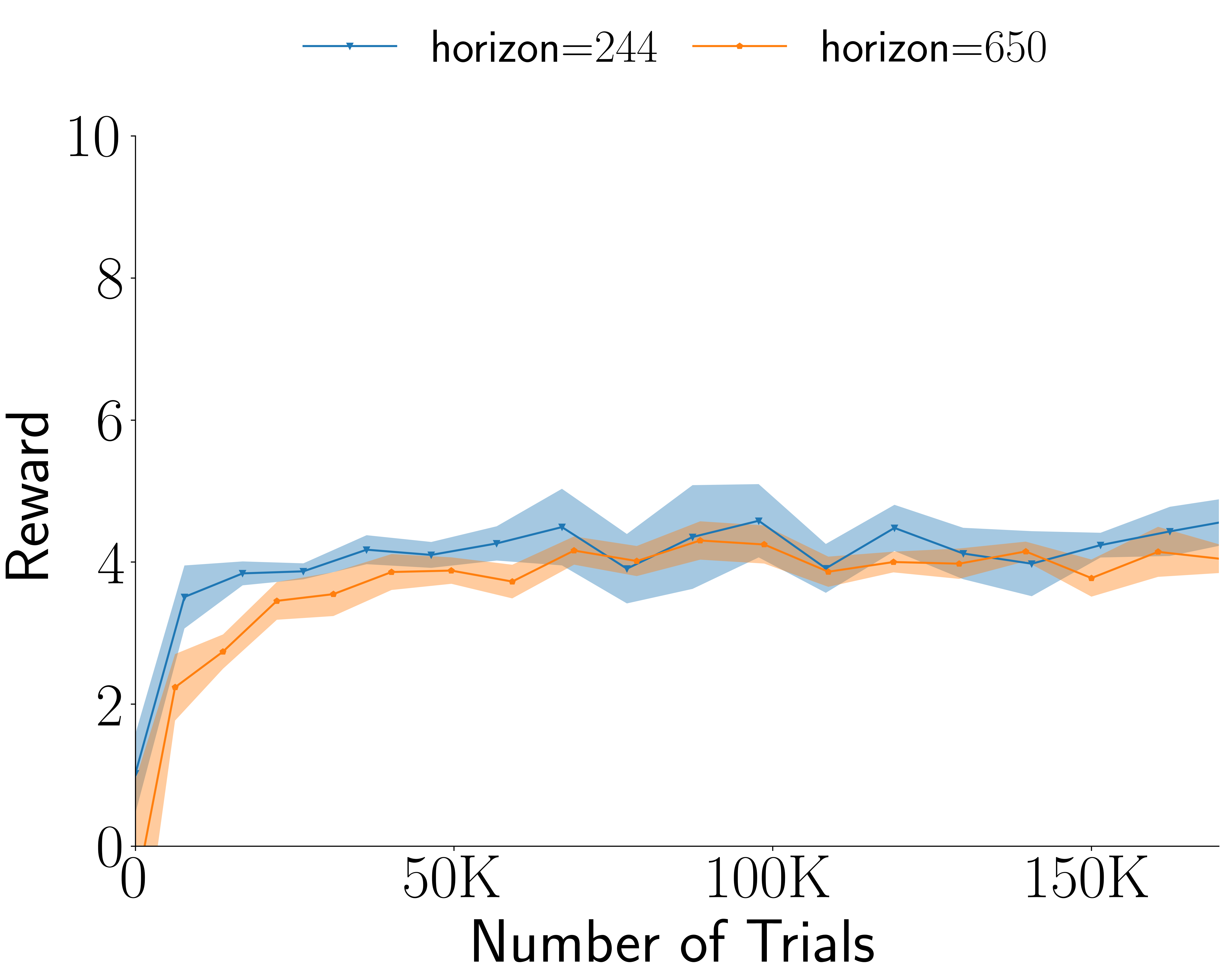}  \\
        \LARGE{a}\\
        \includegraphics[scale=0.23]{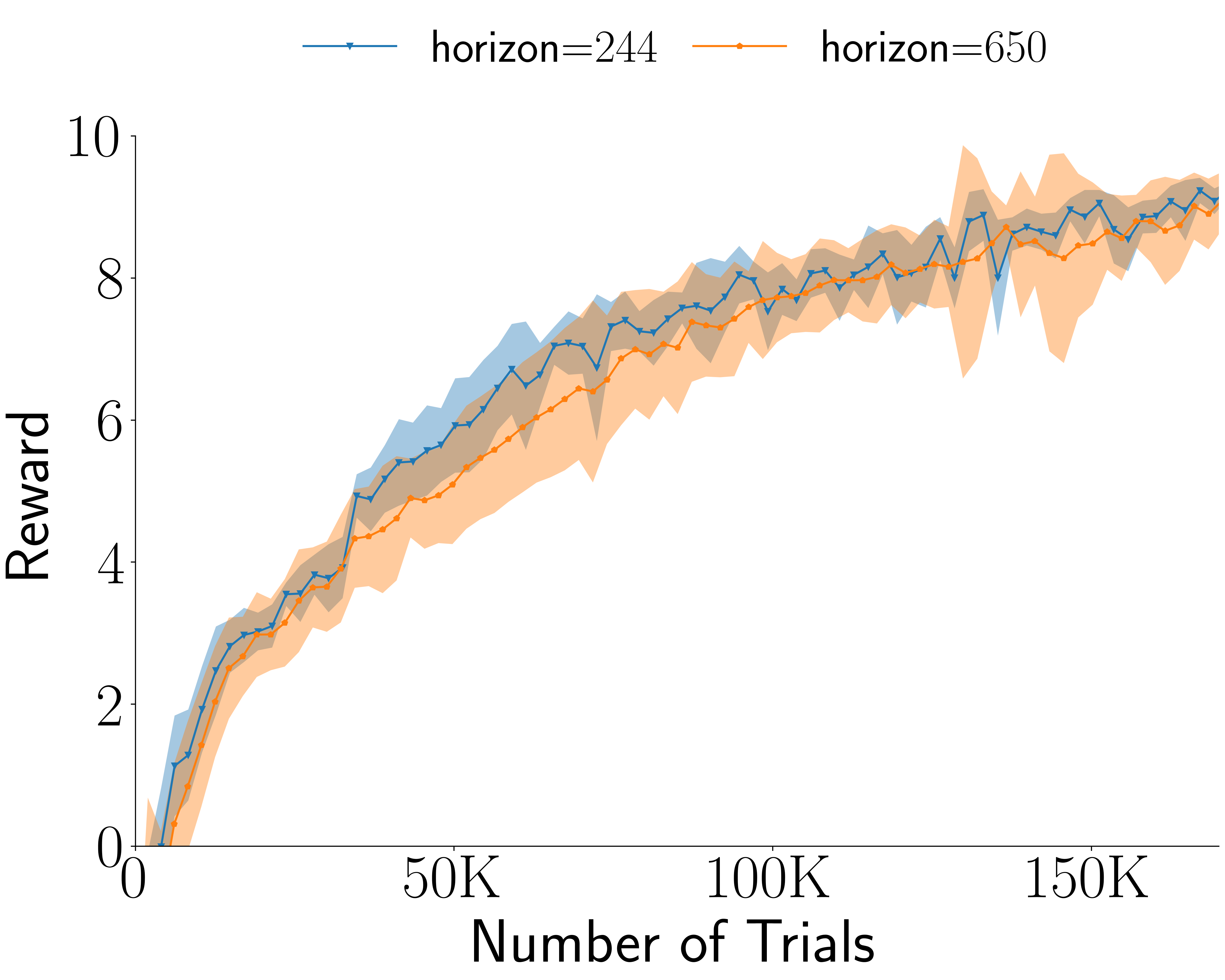} \\
        \LARGE{b}\\
        \end{tabular} 
        \caption{Performances of (a) LSTM and (b) CogRNN  $\tilde{f}$ agents for different horizon lengths.}
        \label{fig:horizon_supplemental}
\end{figure*}

\begin{figure*}[h!] 
    \centering
    \begin{tabular}{c}  
        \includegraphics[scale=0.23]{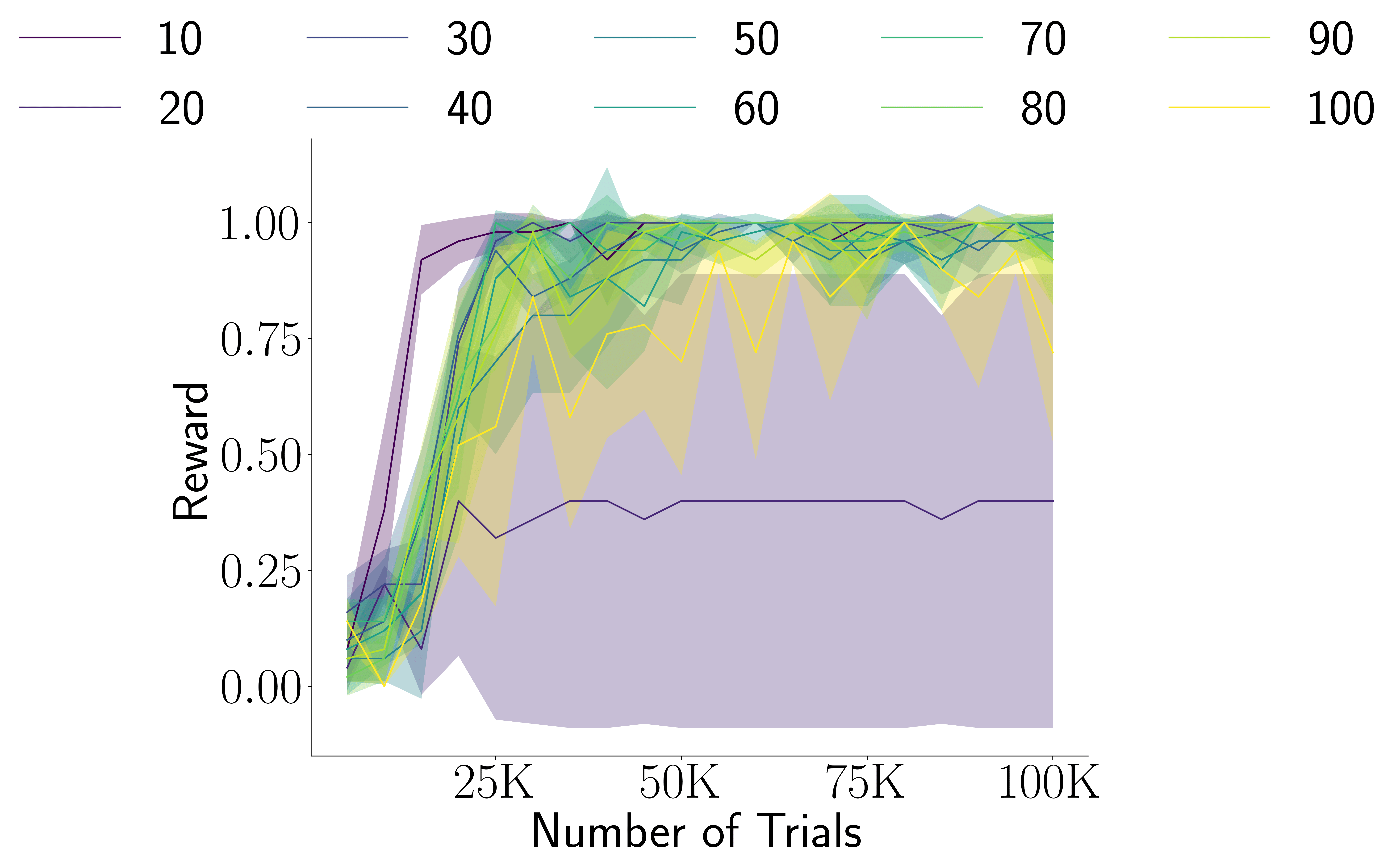}  \\
        \LARGE{a}\\
        \includegraphics[scale=0.23]{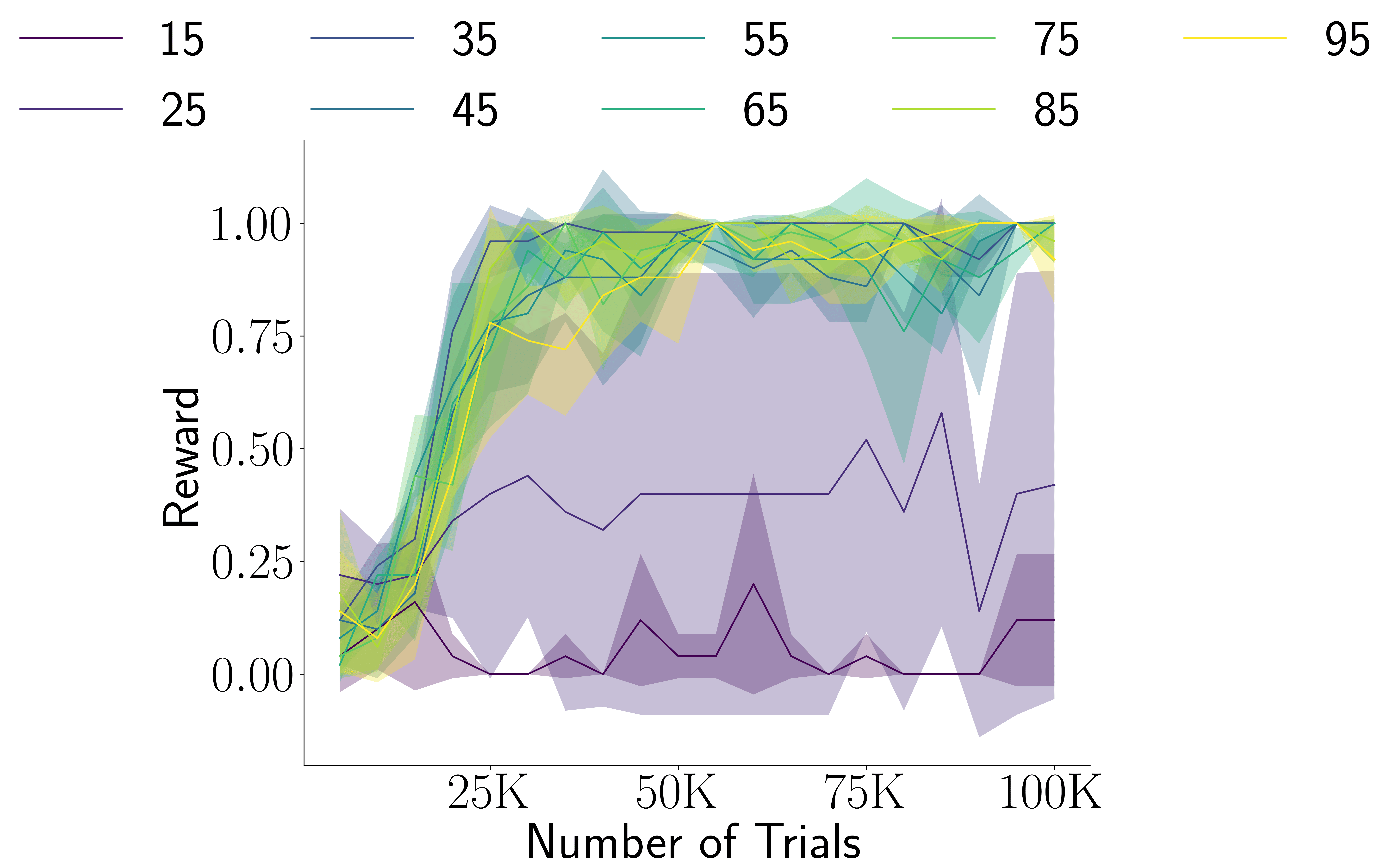} \\
        \LARGE{b}\\
        \end{tabular} 
        \caption{Performances of CogRNN  $\tilde{f}$ agents in learning (a) train and (b) validation intervals for the interval reproduction task.}
        \label{fig:interval_reprod_cogrnn_supplemental}
\end{figure*}

\begin{figure*}[h!] 
    \centering
    \begin{tabular}{c}  
        \includegraphics[scale=0.23]{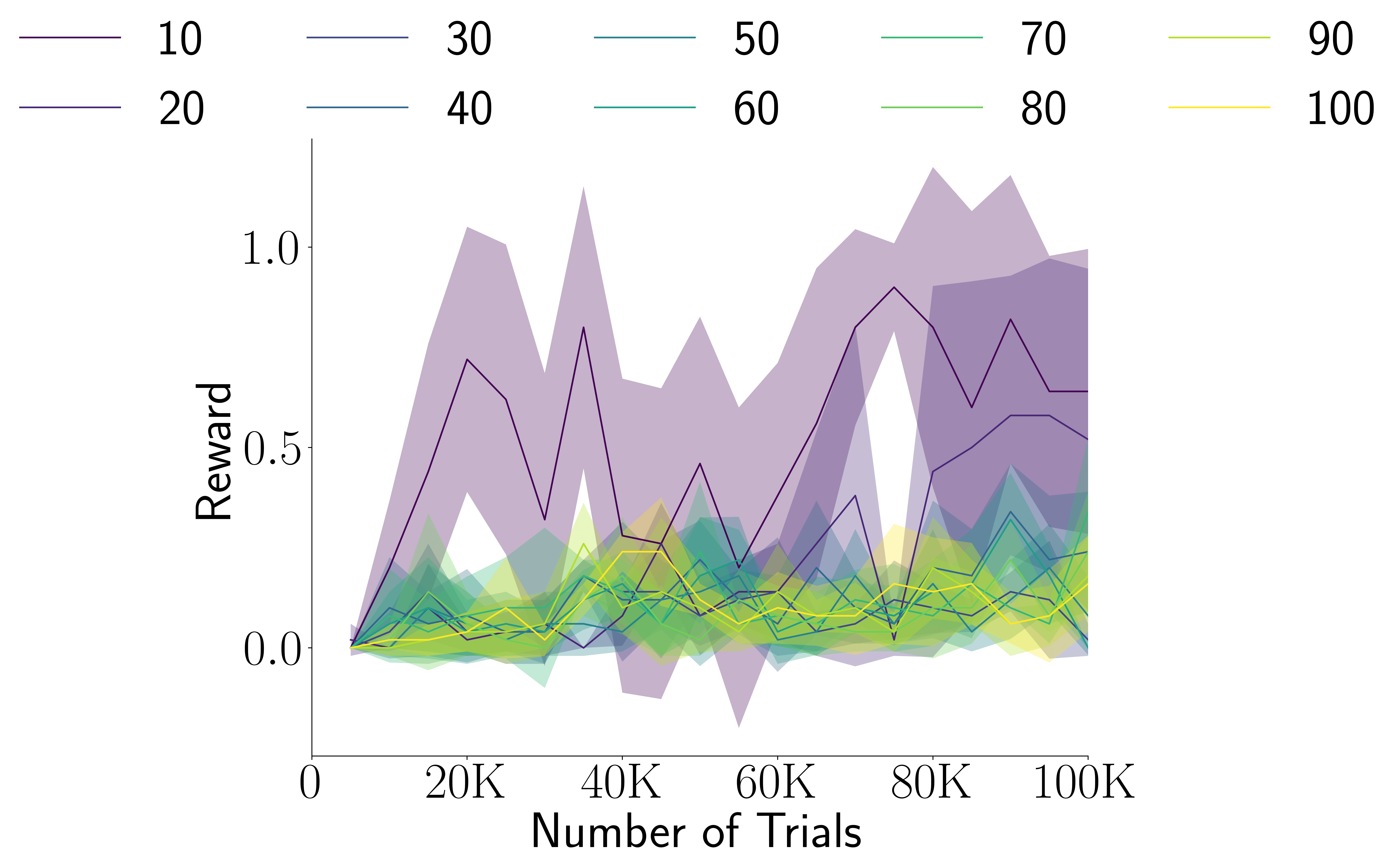}  \\
        \LARGE{a}\\
        \includegraphics[scale=0.23]{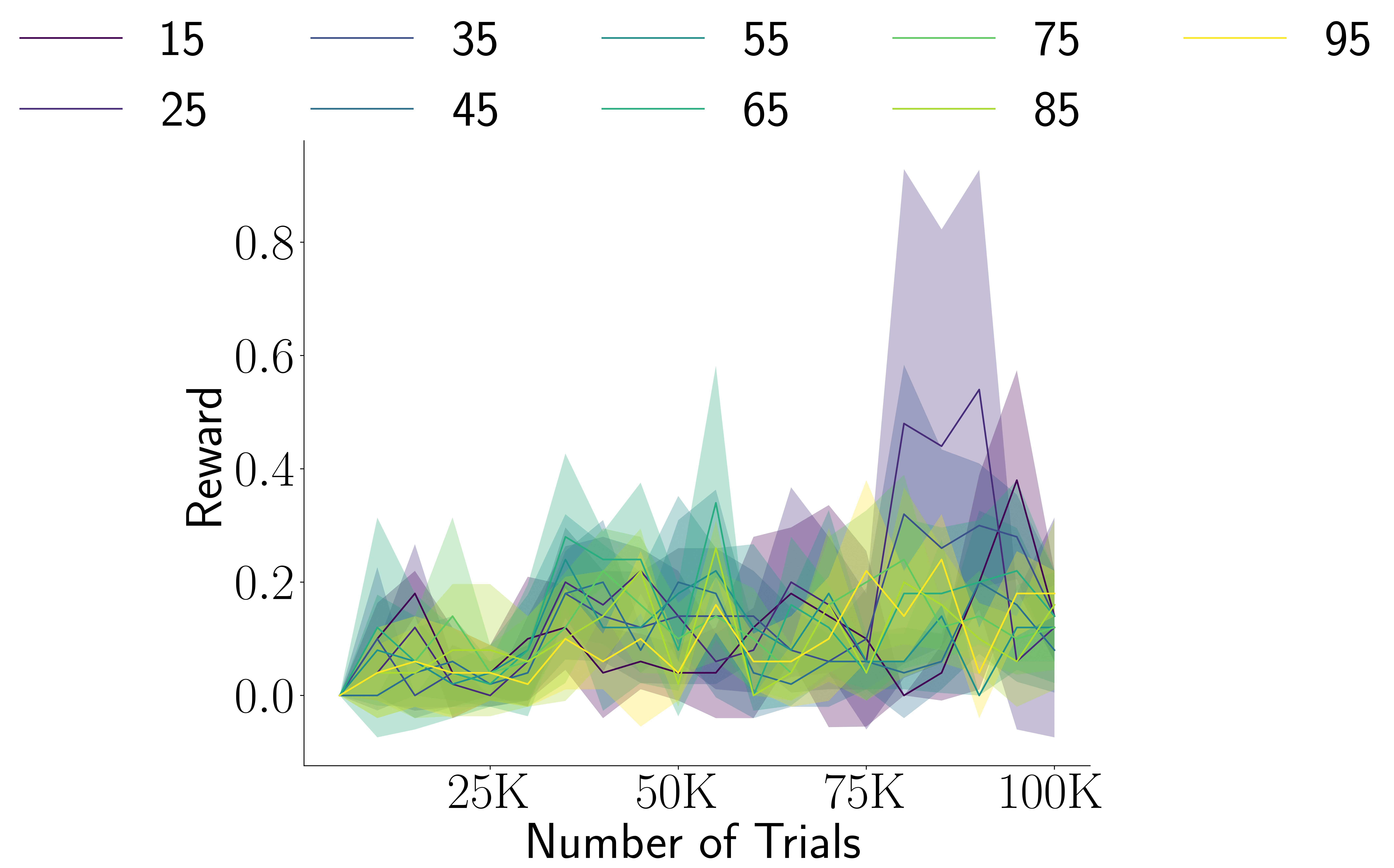} \\
        \LARGE{b}\\
        \end{tabular} 
        \caption{Performances of LSTM  agents in learning (a) train and (b) validation intervals for the interval reproduction task.}
        \label{fig:interval_reprod_rnn_supplemental}
\end{figure*}

As Supplemental Information, we provide additional results on the 3D interval timing environment: 
\begin{itemize}
\item Results of the ablation analysis where agents were trained with frozen RNN, frozen LSTM, and CogRNN $F$ (Fig.~\ref{fig:performance_ablation}). None of the agents reached good performance. 
\item Psychometric curves (Fig.~\ref{fig:psychometric_supplemental}). Similar to rat behavior, agents made the most mistakes for the most difficult time intervals (36 and 40 steps). 
\item Activity heatmaps for neurons with monotonically decreasing/increasing activity (Fig.~\ref{fig:neurons_decay_ramp}). CogRNN $\tilde{f}$ neurons resemble sequentially activated time cells recorded in mammalian brains where width increases with the peak time (Fig.~\ref{fig:peaks_width}).
\item  Results of hyperparameter exploration with different values of:
\begin{itemize}
    \item $\taustar_{min}$  (Fig.~\ref{fig:ablation_f_tilde_tstr_min}),
    \item $\taustar_{max}$ (Fig.~\ref{fig:ablation_f_tilde_tstr_max}),
    \item $k$ (Fig.~\ref{fig:ablation_f_tilde_k}),
    \item learning rate (Fig.~\ref{fig:learning_rate_supplemental}),
    \item entropy coefficient (Fig.~\ref{fig:entropy_supplemental}),
    \item  horizon (Fig.~\ref{fig:horizon_supplemental}).
\end{itemize}
\item Results of training on interval reproduction task (Table~\ref{tab:interval_reprod_perf_CogRNN}, Table~\ref{tab:interval_reprod_perf_LSTM}, Fig.~\ref{fig:interval_reprod_cogrnn_supplemental}, Fig.~\ref{fig:interval_reprod_rnn_supplemental}).
\end{itemize}

\clearpage

\section{Ablation analysis}
\label{sec:performance_ablation}

\section{Psychometric curves}
\label{sec:trained}

\clearpage

\section{Neural activity in RNN, LSTM, and CogRNN agents in 3D interval timing task}
\label{sec:neural_activity_supl}

\clearpage

\section{Hyperparameter exploration for CogRNN in the 3D interval timing task}
\label{sec:params_sweep}

\clearpage

\section{Results of training on Interval Reproduction task}

\begin{table}[h!]
    \centering
    \begin{tabular}{c|cccccccccc}
        \hline
        \textbf{Trial} & \multicolumn{10}{c}{\textbf{Training intervals}} \\ \hline
         & 10 & 20 & 30 & 40 & 50 & 60 & 70 & 80 & 90 & 100 \\ \hline
        25k & .8±.4 & .5±.4 & .5±.4 & .6±.4 & .5±.3 & .8±.2 & .9±.1 & .7±.3 & .6±.2 & .2±.1 \\ 
        50k & .8±.4 & .6±.5 & .8±.4 & .8±.4 & .7±.4 & 1.0±.0 & 1.0±.0 & 1.0±.0 & .9±.1 & .9±.1 \\ 
        100k & 1.0±.0 & .6±.5 & .8±.4 & .8±.4 & 1.0±.0 & 1.0±.0 & 1.0±.0 & .9±.1 & .8±.4 & .6±.4 \\ \hline
    \end{tabular}
    \vspace{0.3cm}

    \begin{tabular}{c|ccccccccc}
        \hline
        \textbf{Trial} & \multicolumn{9}{c}{\textbf{Validation intervals}} \\ \hline
         & 15 & 25 & 35 & 45 & 55 & 65 & 75 & 85 & 95 \\ \hline
        25k & .2±.2 & .5±.4 & .6±.4 & .5±.3 & .6±.3 & .7±.2 & .9±.2 & .6±.2 & .4±.2 \\ 
        50k & .2±.3 & .5±.5 & .8±.4 & .7±.4 & .9±.2 & 1.0±.1 & .9±.1 & 1.0±.1 & .9±.1 \\ 
        100k & .2±.3 & .6±.5 & .7±.4 & .8±.3 & 1.0±.0 & 1.0±.1 & 1.0±.0 & .7±.4 & .8±.4 \\ \hline
    \end{tabular}
    \caption{Mean accuracy and standard-deviation in learning the interval reproduction task for the CogRNN agent.}
    \label{tab:interval_reprod_perf_CogRNN}
\end{table}

\begin{table}[h!]
    \centering
    \begin{tabular}{c|cccccccccc} 
        \hline
        \textbf{Trial} & \multicolumn{9}{c}{\textbf{Training intervals}} \\ \hline
         & 10 & 20 & 30 & 40 & 50 & 60 & 70 & 80 & 90 & 100 \\ \hline
        25k & 1.0±.0 & .1±.1 & .1±.1 & .2±.1 & .2±.1 & .1±.0 & .1±.0 & .0±.1 & .1±.1 & .1±.1 \\
        50k & .4±.4 & .2±.2 & .2±.2 & .1±.1 & .1±.1 & .3±.1 & .3±.1 & .1±.1 & .1±.1 & .1±.1 \\
        100k & 1.0±.0 & .4±.4 & .4±.3 & .2±.1 & .2±.1 & .2±.1 & .2±.1 & .1±.1 & .1±.1 & .1±.1 \\ \hline
    \end{tabular}
    \vspace{0.3cm}

    \begin{tabular}{c|ccccccccc}
    \hline
        \textbf{Trial} & \multicolumn{9}{c}{\textbf{Validation intervals}} \\ \hline
         & 15 & 25 & 35 & 45 & 55 & 65 & 75 & 85 & 95 \\ \hline
        25k & .2±.2 & .1±.1 & .1±.0 & .2±.1 & .2±.1 & .1±.1 & .2±.2 & .2±.2 & .3±.1 \\ 
        50k & .1±.1 & .2±.3 & .2±.1 & .1±.1 & .1±.1 & .2±.1 & .2±.1 & .1±.0 & .1±.1 \\ 
        100k & .6±.4 & .3±.4 & .3±.2 & .2±.1 & .1±.1 & .1±.1 & .1±.0 & .1±.1 & .1±.1 \\ \hline
    \end{tabular}
    \caption{Mean accuracy and standard-deviation in learning the interval reproduction task for the LSTM agent.}
    \label{tab:interval_reprod_perf_LSTM}
\end{table}

\end{document}